%% file: main.tex
\documentclass[10pt,twocolumn,letterpaper]{article}

\usepackage[pagenumbers]{cvpr} % To force page numbers, e.g. for an arXiv version

\input{preamble}

\usepackage{multirow}
\usepackage{color, colortbl}
\usepackage{pbox}
\usepackage{makecell}
\usepackage{caption}
\usepackage{graphicx}
\usepackage{booktabs}
\usepackage{pifont}
\definecolor{cvprblue}{rgb}{0.21,0.49,0.74}
\usepackage[pagebackref,breaklinks,colorlinks,allcolors=cvprblue]{hyperref}

\title{Med-MMFL: A Multimodal Federated Learning Benchmark in Healthcare}

\author{Aavash Chhetri$^2$\footnotemark[1] \quad Bibek Niroula$^2$\footnotemark[1] \quad Pratik Shrestha$^2$ \quad Yash Raj Shrestha$^3$ \quad Lesley A Anderson$^1$\\
Prashnna K Gyawali$^4$ \quad Loris Bazzani$^5$ \quad Binod Bhattarai$^{1,2,6}$\footnotemark[2]\\ 
\\
$^1$University of Aberdeen, Aberdeen, UK\\
 $^2$NepAl Applied Mathematics and Informatics Institute for research, Nepal\\$^3$University of Lausanne, Switzerland\\$^4$West Virginia University, USA\\$^5$University of Verona, Italy\\$^6$ University College London, UK
}
\begin{document}

\maketitle
\renewcommand{\thefootnote}{\fnsymbol{footnote}} 
\footnotetext[1]{Equal Contribution.} 
\footnotetext[2]{Binod Bhattarai is the corresponding author.}
\input{sec/0_abstract}
\input{sec/1_intro}

\input{sec/2_relatedworks}

\input{sec/3_datasets}
\input{sec/4_method}

\input{sec/5_experiments}

\input{sec/6_conclusion}

\section*{Acknowledgements}
This work was supported as part of the “Swiss AI initiative” by a grant from the Swiss National Supercomputing Centre (CSCS) under project ID a168 on Alps.

% \clearpage
{
    \small
    \bibliographystyle{ieeenat_fullname}
    \bibliography{main}
}

\input{sec/X_suppl}

\end{document}

%% file: preamble.tex
\usepackage{xcolor}
\usepackage{colortbl}

%% file: sec/0_abstract.tex
\begin{abstract}
Federated learning (FL) enables collaborative model training across decentralized medical institutions while preserving data privacy. However, medical FL benchmarks remain scarce, with existing efforts focusing mainly on unimodal or bimodal modalities and a limited range of
medical tasks. This gap underscores the need for standardized evaluation to advance systematic understanding in medical MultiModal FL (MMFL). To this end, we introduce Med-MMFL, the first comprehensive MMFL benchmark for the medical domain, encompassing diverse modalities, tasks, and federation 
scenarios. Our benchmark evaluates six representative state-of-the-art FL algorithms, covering different aggregation strategies, loss formulations, and regularization techniques. It spans datasets with 2 to 4 modalities, comprising a total of 10 unique medical modalities, including text, pathology images,
ECG, X-ray, radiology reports, and multiple MRI sequences. Experiments are conducted across naturally federated, synthetic IID, and synthetic non-IID settings to simulate real-world heterogeneity. We assess segmentation, classification, modality alignment (retrieval), and VQA tasks. To support reproducibility and fair comparison of future multimodal federated learning (MMFL) methods under realistic medical settings, we release the complete benchmark implementation, including data processing and partitioning pipelines, at \url{https://github.com/bhattarailab/Med-MMFL-Benchmark}.

\end{abstract}

%% file: sec/1_intro.tex
\section{Introduction}
\label{sec:intro} 
%% WHY MULTIMODAL?
Clinicians inherently rely on information from multiple sources and modalities to perform reliable diagnosis, prognosis, and formulating treatment plans. 
This reliance on heterogeneous information has motivated the recent development of multimodal models for healthcare applications~\cite{Acosta2022Multimodal, KRONES2025102690, jmirmultimodal}. Like clinicians, the goal of multimodal models is to provide an holistic view of the disease and offer improved and consistent diagnostic performance~\cite{Venugopalan2021Multimodal, Shrestha2023Medical} by integrating information across diverse data sources like medical scans, omics data, and pathology reports.
Training such multimodal models in healthcare is challenging because of the fragmentation of medical data owing to their sensitive nature and strict privacy concerns that limit their broad distribution.
%% WHY FEDERATED LEARNING?
Federated learning (FL) addresses these challenges by enabling privacy-preserving training of local models for each institution (clients), while aggregating these models at a global level (server) without any exchange or distribution of patient data~\cite{McMahan2017Communication}.
More recently, numerous unimodal (e.g., imaging-only)  and bimodal (e.g., image–text) FL methods~\cite{Xu2021Federated, GUAN2024110424unifedmed, KRONES2025102690, jmirmultimodal} have been proposed to deal with these challenges. 
However, it still remains open how to scale beyond two modalities and there is a lack of standardized evaluation for different methods in a reproducible manner.
To address these limitations, we present a comprehensive benchmark which is composed by multiple modalities, datasets, tasks, evaluation protocols, and baselines with the aim to promote reproducibility and fair comparison and to facilitate fast progress in the domain of multimodal healthcare.

\begin{table*}[t]
  \centering
  \newcommand{\y}{\ding{51}}
  \newcommand{\n}{\ding{55}}
  \small % slightly larger than \footnotesize
  \setlength{\tabcolsep}{4.5pt} % reduce column spacing
  \renewcommand{\arraystretch}{1.1} % tighten vertical spacing
  \resizebox{0.95\linewidth}{!}{%ddcfdsf
    \begin{tabular}{|c|c|>{\columncolor{gray!10}}c|>{\columncolor{gray!10}}c|>{\columncolor{gray!10}}c|>{\columncolor{blue!10}}c|>{\columncolor{blue!10}}c|>{\columncolor{green!10}}c|}
      \hline
        \multicolumn{2}{|c|}{Features} & NIID-Bench \cite{li2022federated} & FLamby \cite{NEURIPS2022_232eee8e}  & FedLLM-Bench \cite{NEURIPS2024_c8cdab0e} & FedMultimodal \cite{Feng2023FedMultimodal} & FedVLMBench \cite{zheng2025fedvlmbenchbenchmarkingfederatedfinetuning} & \textbf{Med-MMFL (ours)} \\ \hline \hline
        \multirow{2}{*}{Modalities}
        & \# Modalities (min, max) & (1,1) & (1,1) & (1,1) & \underline{(2,2)}  & \underline{(2,2)} & \textbf{(2,4)} \\ \cline{2-8}
        & \# Unique Medical Modalities & 0 & \underline{5} & 0 & 2  & 2 & \textbf{10} \\ \hline
        \multicolumn{2}{|c|}{\# Multimodal Medical Datasets} & 0 & 0 & 0 & 1 & \underline{2} & \textbf{5} \\ \hline
        \multicolumn{2}{|c|}{\# Distinct FL Algorithms} & \underline{4} & \underline{4} & \underline{4} & \underline{4} & 3 & \textbf{6} \\ \hline
        \multirow{3}{*}{Partitioning Strategies} & Real-world & \y & \y & \y & \y & \n & \y \\ \cline{2-8} 
        & Synthetic IID & \y & \n & \y & \y & \y & \y \\ \cline{2-8} 
        & Synthetic non-IID & \y &  \n & \n & \y & \y & \y \\ \hline
        \multicolumn{2}{|c|}{Evaluation Tasks} & 1 & \underline{3} & 2 & 1 & \textbf{4}& \textbf{4} \\ \hline
    \end{tabular}%
      }
  \captionsetup{width=0.95\linewidth}
  % \caption{Experimental settings in existing federated medical, multimodal and/or language model benchmarks and our benchmark Med-MMFL. For each feature, \textbf{bold} highlights the most comprehensive benchmark, while \underline{underlined} indicates the second best.}
\vspace{-2mm}
\caption{Comparison of experimental settings across unimodal benchmarks 
(shaded in \colorbox{gray!10}{\strut grey}), multimodal benchmarks 
(shaded in \colorbox{blue!10}{\strut blue}), and our Med-MMFL benchmark 
(shaded in \colorbox{green!10}{\strut green}). For each feature, 
\textbf{bold} indicates the most comprehensive support and 
\underline{underlined} indicates the second best.}
\vspace{-4mm}
  \label{tab:benchmarks-comparison}
\end{table*}

% IDENTIFYING GAPS IN EXISTING BENCHMARKS
To facilitate FL research, several FL benchmarks have been proposed~\cite{caldas2019leafbenchmarkfederatedsettings, he2020fedmlresearchlibrarybenchmark, li2022federated, lai2022fedscalebenchmarkingmodelperformance}. 
However, they remain fragmented across domains and objectives. NIID-Bench~\cite{li2022federated} focuses on algorithmic diversity under controlled non-IID settings but is limited to unimodal data. FLamby~\cite{NEURIPS2022_232eee8e} provides the first medical FL suite, yet without multimodality. In contrast, existing multimodal FL benchmarks such as FedMultimodal~\cite{Feng2023FedMultimodal}, FedVLMBench~\cite{zheng2025fedvlmbenchbenchmarkingfederatedfinetuning}, and FedLLM-Bench~\cite{NEURIPS2024_c8cdab0e} expand to multiple modalities, yet there remains no comprehensive benchmark that jointly captures the multimodal, medical, and federated aspects of real-world healthcare scenarios. 

A closer examination of existing benchmarks (see ~\cref{tab:benchmarks-comparison}) reveals several key limitations. \textbf{(1) Datasets:} Existing medical FL benchmark, such as~\cite{NEURIPS2022_232eee8e} focus on unimodal datasets, where as multimodal benchmarks~\cite{zheng2025fedvlmbenchbenchmarkingfederatedfinetuning, Feng2023FedMultimodal} largely ignore medical data. \textbf{(2) Modalities: } Prior multimodal works \cite{Feng2023FedMultimodal, zheng2025fedvlmbenchbenchmarkingfederatedfinetuning} are typically limited to two modalities per dataset, and hence, offer only a limited insight into the characteristics of real-world multimodal clinical data in federated settings. \textbf{(3) Tasks:} Most benchmarks cover only a handful of task types (\cref{tab:benchmarks-comparison}), highlighting the need for broader and more diverse evaluations of medical tasks. \textbf{(4) Partitioning:} Partitioning strategies in existing benchmarks remain restricted in scope:~\cite{NEURIPS2022_232eee8e, NEURIPS2024_c8cdab0e} focus only on real-world splits, whereas~\cite{zheng2025fedvlmbenchbenchmarkingfederatedfinetuning} employs synthetic ones, leaving few frameworks that combine both realistic hospital-level silos and  synthetic non-IID settings required for comprehensive FL evaluation. \textbf{(5) Algorithms coverage:} The set of FL Algorithms studied in existing benchmarks remains narrow, with many variants (e.g., FedAvgM \cite{49350hsu}, FedAdam, FedAdagrad, FedYogi) reducible to the unified FedOpt formulation~\cite{50448reddi}. Consequently, existing evaluations offer limited exploration of fundamentally distinct state-of-the-art FL algorithms, urging the need for a broader and more inclusive benchmark.

\begin{figure*}[t]
    \centering
    \includegraphics[width=\linewidth]{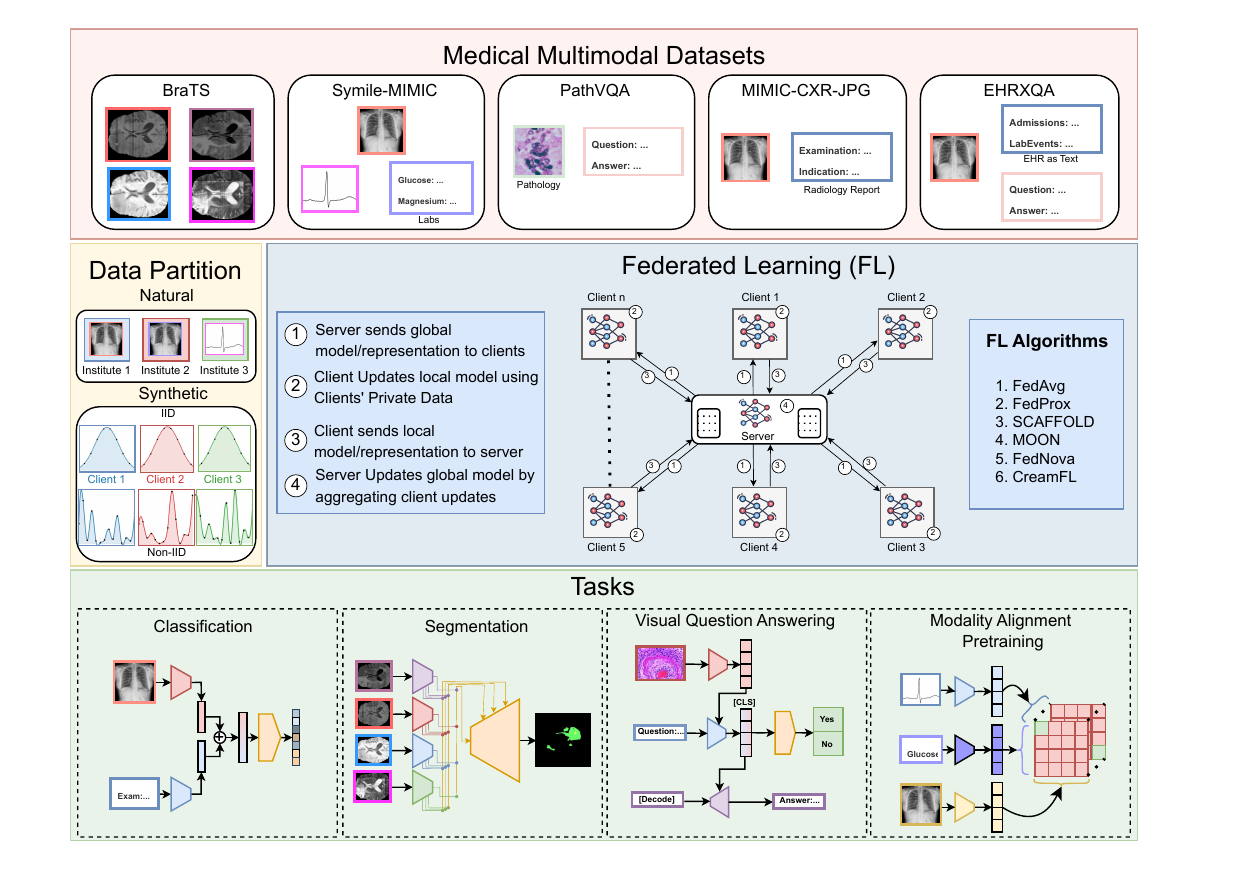}

    \captionsetup{width=0.9\linewidth}
    \vspace{-4mm}

    \caption{Overview of our proposed Med-MMFL benchmark framework. It spans diverse multimodal medical datasets, task types, and client partitioning strategies, integrating multiple FL algorithms to provide a unified evaluation platform.}
    \label{fig:med-mmfl-framework}
    \vspace{-4mm}
\end{figure*}

%Proposing our Benchmark
To address theses limitations, we introduce \textbf{Med-MMFL} (~\cref{fig:med-mmfl-framework}), a comprehensive standardized benchmark designed specifically for multimodal federated learning in healthcare. Our main contributions can be summarized as follows:

\begin{enumerate}
    \item Med-MMFL integrates \textbf{6} distinct FL Algorithms, \textbf{4} types of tasks, \textbf{3} data partitioning strategies, and  \textbf{5} medical datasets with varying degree of multimodality (\textbf{2 to 4 modalities}).
    \item We generalize existing algorithms (e.g., MOON, CreamFL) to handle more than two modalities for fair multimodal evaluation.
    \item To foster transparency and facilitate reproducible research, we publicly release our benchmark implementation, including all the dataset processing and partitioning pipelines.
\end{enumerate}

%Paper Organization
The remainder of this paper is organized as follows.~\cref{sec:related-works} reviews related work on federated learning and existing benchmarks.~\cref{sec:datasets} describes the Med-MMFL datasets, including data partitioning and baseline setups.~\cref{sec:method} presents the Med-MMFL framework and the adapted FL algorithms. Finally,~\cref{sec:experiments} describes the experimental setup and presents the benchmark results.

%% file: sec/2_relatedworks.tex
\section{Related Works}
\label{sec:related-works}
 
\paragraph{FL algorithms.}
%% FedAvg based methods
Most of modern FL approaches are based on FedAvg~\cite{McMahan2017Communication}, a foundational work which proposed to train a global model by averaging the models trained locally by each party (clients) while keeping their data private. Subsequent studies~\cite{li2020convergencefedavgnoniiddata, 49350hsu, li2022federated} built on FedAvg establish that non-Independently and Identically Distributed (non-IID) data degrade the convergence rate of FedAvg. To address this issue of statistical heterogeneity, FedProx \cite{li2020federatedprox} adds a $L_2$ regularizer to the loss function that restricts local updates to be close to the global model. 
SCAFFOLD~\cite{karimireddy2021scaffoldstochasticcontrolledaveraging} introduces control variates to correct the drift of local updates, ensuring their directions remain consistent with the global optimization path.
FedNova \cite{wang2020tacklingobjectiveinconsistencyproblemnova} corrects FedAvg’s bias toward clients with more local updates by normalizing and scaling their contributions during the aggregation stage.
MOON \cite{li2021modelcontrastivefederatedlearningMOON} addresses the issue of non-IID data via
 contrastive learning in model-level by comparing the representations to bridge the gap between the representations learned by the local model and the global model. FedDyn \cite{acar2021federatedlearningbaseddynamicfeddyn} dynamically adjusts each client’s objective with a regularization term, ensuring convergence toward stationary points of the global empirical loss. 
 % distillation based methods
 To address the constraint of identical client architectures in all these FedAvg-derived algorithms, knowledge-distillation-based fusion methods \cite{lin2021ensembledistillationrobustmodelFedDF, Wu_2022FedKD, cho2022heterogeneousensembleknowledgetransferFedET} have been proposed to enable heterogeneous participation by replacing parameter aggregation with model-agnostic knowledge transfer. CreamFL \cite{yu2023multimodalfederatedlearningcontrastiveCreamFL} advances knowledge-distillation fusion toward multimodal learning by adopting representation-level transfer and contrastive objectives to reduce model drift. 
 %scope of our work
 Although our benchmark focuses on a representative set of fundamentally distinct FL algorithms, some state-of-the-art methods~\cite{acar2021federatedlearningbaseddynamicfeddyn, li2021fedbnfederatedlearningnoniid, 50448reddi} explore complementary approaches and are not directly evaluated in this work. Furthermore, an active and promising research direction of Personalized FL \cite{Tan_2023PersonalizedFed, dai2024federatedmodalityspecificencodersmultimodaFEDMemal} lies outside the scope of our current work and is open for future exploration. 
\vspace{-2mm}

\paragraph{FL benchmarks.}
LEAF \cite{caldas2019leafbenchmarkfederatedsettings} was an early benchmark providing federated datasets with realistic natural splits. FedML~\cite{he2020fedmlresearchlibrarybenchmark} offers an open research library and benchmark to facilitate FL algorithm development and fair performance comparison.
NIID-Bench \cite{li2022federated} proposes diverse non-IID partitioning strategies and evaluates multiple FL algorithms, forming a comprehensive unimodal benchmark. However, these benchmarks focus on unimodal and non-medical data, leaving a gap for multimodal medical FL evaluation. FLamby \cite{NEURIPS2022_232eee8e} marked a major milestone in FL for healthcare research by targeted benchmark  with seven medical datasets covering five distinct input modalities. Nevertheless, its unimodal and real-world partitioning design, along with limited representative FL algorithms leaves open opportunities for the community to explore broader multimodal and data heterogeneity scenarios. FedMultimodal \cite{Feng2023FedMultimodal} introduced a FL benchmark targeting diverse multimodal applications. However, it is limited by its focus on cross-device settings, single evaluation task coverage, as well as datasets with only two modalities. Consequently, the evaluation of federated learning algorithms across scenarios involving more than two modalities remains largely unexplored. Furthermore, its sole healthcare data PTBXL \cite{Wagner2020ptbxl} is not inherently a multimodal dataset. While the ECG signals are split and passed as ``separate modalities", they all stem from the same underlying modality, merely grouped across electrode groups. FedLLM-Bench \cite{NEURIPS2024_c8cdab0e} establishes a federated benchmark for LLM training (instruction tuning and preference alignment). Building upon it, FedVLMBench \cite{zheng2025fedvlmbenchbenchmarkingfederatedfinetuning} introduces a systematic benchmark for federated fine-tuning of Vision-Language Models (VLMs) with three FL algorithms, while incorporating two medical datasets. Despite its contributions, the benchmark still falls short in several aspects: its medical datasets are limited in diversity, the evaluation of FL algorithms remains narrow, and it addresses only two modalities (image and text). Thus, despite prior work, a benchmark that fully integrates multimodality, medical-data diversity, and federated realism remains absent. Thus, we propose Med-MMFL, providing comprehensive coverage of datasets, modalities, partitioning strategies, and FL algorithms for systematic evaluation.

%% file: sec/3_datasets.tex
\section{Med-MMFL Datasets}
\label{sec:datasets}
In this benchmark, we carefully select five publicly available medical datasets covering diverse modalities and clinical tasks, chosen for their widespread adoption and/or representativeness of real-world medical challenges. Since these datasets are not inherently federated, we systematically transform them into FL-compatible versions through realistic client partitioning and adapting baselines to capture broader and clinically meaningful task diversity.

\begin{figure}[h]
    \centering
    \includegraphics[width=0.9\linewidth]{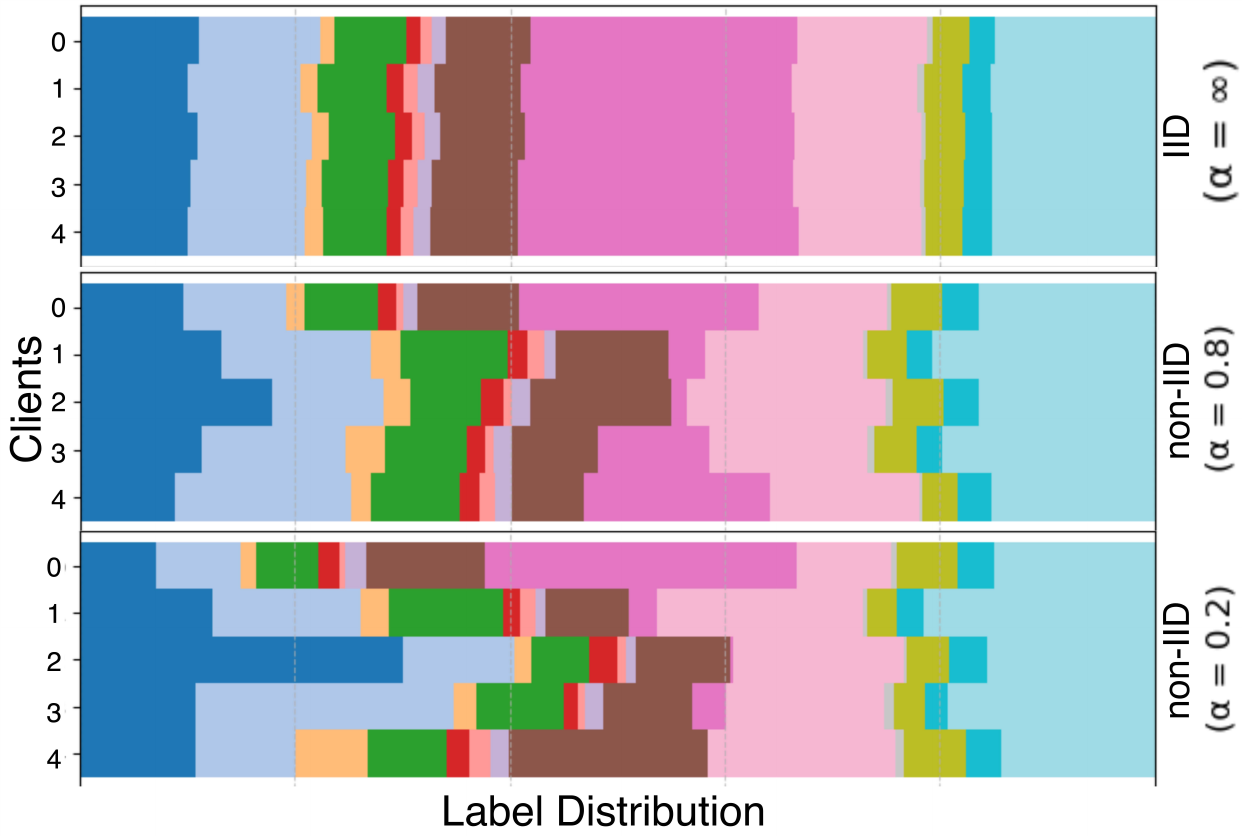}
    
    \caption{Representative client-level label distribution for the MIMIC-CXR-JPG dataset obtained using our federated partitioning strategy. IID splits produce similar label frequencies across clients, whereas non-IID splits yield heterogeneous distributions where certain labels dominate or are absent on specific clients. Other datasets exhibit analogous distribution patterns under the same protocol (see~\cref{sec:supp_datasets})}
    \label{fig:label_dist}
    \vspace{-4mm}
\end{figure}

\subsection{Fed-BraTS-GLI2024}
\label{subsec:brats24}
BraTS-GLI2024~\cite{deverdier20242024braintumorsegmentation, karargyris2023federated} is a multimodal 3D Magnetic
resonance imaging (MRI) dataset for Brain Tumor Segmentation with four MRI modalities: 1. Pre-contrast T1-weighted (T1), 2. Contrast-enhanced T1-weighted (T1-Gd), 3. T2-weighted (T2), and 4. T2-weighted fluid-attenuated inversion recovery (FLAIR), along with the segmentation mask for each patient. We build a federated version of BraTS-GLI2024 with \textbf{5} clients, each corresponding to a unique contributing center in the dataset, and we refer to this setup as a \textit{natural} partition. Additionally, we use pseudo-classes for \textit{synthetic} data partitioning. Specifically, we apply clustering on the class proportion vectors, computed as the normalized voxel counts per class for each segmentation mask, and treat each cluster as a pseudo-class. To simulate an IID scenario, we uniformly distribute the samples across clients such that each client receives similar proportions of all pseudo-classes. For non-IID synthetic partitions, we follow the de-facto approach from~\cite{yurochkin2019bayesiannonparametricfederatedlearning} used in many works~\cite{li2022federated, 49350hsu, Feng2023FedMultimodal, zheng2025fedvlmbenchbenchmarkingfederatedfinetuning}, which is based on sampling from a Dirichlet distribution.
Specifically, we sample $p_k \sim Dir_N(\alpha)$ and allocate a $p_{k,j}$ proportion of the instances of pseudo-class $k$ to party $j$. Here, $Dir(\cdot)$ denotes the Dirichlet distribution and $\alpha$ is the concentration parameter which we use to vary the degree of pseudo-class skew across clients, thereby modulating the level of data heterogeneity in the federated setup.
As a baseline for the segmentation task, we train the RFNet \cite{ding2021rfnet} architecture following its reference implementation. The experiments are evaluated using Dice Score Coefficient (DSC).

\subsection{Fed-MIMIC-CXR-JPG}
The MIMIC-CXR-JPG dataset~\cite{johnson2019mimiccxrjpglargepubliclyavailable}, derived from MIMIC-CXR \cite{Johnson2019, goldberger2000physionet, johnson2024mimiccxr}, pairs Chest X-ray (CXR) images with their corresponding textual radiology reports. The dataset provides 14 labels that correspond to radiology findings to form a multi-label classification task measured by the macro Area Under the ROC Curve (AUC). Like ~\cref{subsec:brats24}, we use Dirichlet distribution based sampling to generate \textit{synthetic} federated partitions,~\cref{fig:label_dist}. To extend label-distribution skew~\cite{li2022federated} to multi-label dataset, each label dimension is treated independently: for every class, the corresponding samples are allocated to clients according to proportions drawn from a Dirichlet distribution. In the multi-label scenario, a sample may belong to multiple classes. Hence, when overlap occurs, the final client assignment of such samples effectively follows the allocation determined by the last class processed. To ensure homogeneous distribution in IID partitioning, we use high value ($\alpha \to \infty $) of the concentration parameter $\alpha$ while sampling from the Dirichlet distribution. As a baseline, we train a multimodal classifier following the implementation of \cite{poudel2024carmflcrossmodalaugmentationretrieval}.

\subsection{Fed-Symile-MIMIC}
Symile-MIMIC \cite{saporta2025symilemimic, saporta2024contrastingsymilesimplemodelagnostic, goldberger2000physionet} is a multimodal clinical dataset comprising chest X-rays (CXR), electrocardiograms (ECG), and blood laboratory measurements (blood labs). Synthetic federated partitions are derived using the patient and admission metadata. To induce controllable non-IID characteristics, we sample client-wise proportions from a Dirichlet distribution, parameterized by the empirical distribution of the metadata values. 
Since geographical information of the centers was not available, this partitioning is \textit{synthetic}, nonetheless the most realistic option with this dataset.
Our baseline model follows~\cite{saporta2024contrastingsymilesimplemodelagnostic}, which performs modality alignment pretraining and we evaluate the results through a zero-shot retrieval setup, wherein CXRs are retrieved based on corresponding ECG and lab representations. We use the ResNet-50
and ResNet-18 architectures \cite{7780459resnet} for the CXR and ECG encoders, respectively, and a three-layer Multi-layer Perceptron (MLP) to encode the blood labs data following the implementation of \cite{saporta2024contrastingsymilesimplemodelagnostic}. The downstream retrieval task is evaluated using the Accuracy metric. 

\subsection{Fed-PathVQA}
\label{subsec:fpvqa}
PathVQA \cite{he2020pathvqa30000questionsmedical} is a dataset for pathological Visual Question Answering (PVQA) designed to emulate diagnostic reasoning similar to that assessed in the American Board of Pathology examinations. The dataset comprises 4,998 pathology images and 23,700 question-answer pairs. For our experiments, we formulate a close-ended Visual Question Answering (VQA) task by using the yes/no subset of the dataset, which includes 16,334 VQA instances (49.8\% of the total).
We construct pseudo-classes for partitioning, as described in~\cref{subsec:brats24}, distribute them uniformly across clients and use Dirichlet sampling to generate IID and non-IID \textit{synthetic} partitions. 
To infer pseudo-classes, we perform clustering of questions embedded using BioMedClip~\cite{zhang2024biomedclip}, a powerful text model for biomedical applications.
As a baseline, we train BLIP \cite{blip} and pass the [CLS] token from the text encoder into a 2 layer MLP for binary (yes/no) classification. The quality of the resulting close-ended VQA task is measured by the F1 Score.

\subsection{Fed-EHRXQA}
EHRXQA \cite{bae2023ehrxqamultimodalquestionanswering} is a multi-modal question-answering dataset which includes structured Electronic Health Records (EHRs) and chest X-ray images. The original dataset uses SQL queries over the full EHR database to retrieve patient-specific or multi-patient information. In practice, however, clinical reasoning occurs at the individual patient level, where clinicians interpret multimodal data to answer patient-specific questions. Thus, database-wide SQL retrieval is redundant. By converting each patient’s structured EHR into text and removing the SQL module, we simplify the VQA task to focus on individual patient data. This preprocessing yields 9,956 QA pairs, each aligned with a single patient record. Consistent with  ~\cref{subsec:fpvqa}, we split the dataset into synthetic IID and non-IID partitions by generating embeddings via BioMedClip \cite{zhang2024biomedclip} and distribute samples based on the clusters of the embeddings. As a baseline, we train the same architecture of BLIP \cite{blip}, specifically with its generative decoder to generate open-ended answers. The resulting VQA task is evaluated by Token Overlap F1 score \cite{deyoung-etal-2020-eraser}.

%% file: sec/4_method.tex
\section{The Med-MMFL Benchmark Framework}
\label{sec:method}

We consider a multimodal federated learning setting with $C$ clients, each having its private dataset $D_C$ with $n_C$ samples. The $i^{th}$ data sample in $D_C$ is represented by tuple $(\{X_m^{(i)}\}_{m=1}^{M_C}, Y^{(i)})$, where $Y^{(i)}$ and $M_C$ represent the label set and the number of modalities in the $C^{th}$ client, respectively. The clients collaboratively train a global model 
$f_s(\cdot; \mathbf{w}): \mathbb{R}^{n} \rightarrow \mathbb{R}^{d}$ 
parameterized by $\mathbf{w}$, which maps inputs to outputs of dimension $d$.
Note that while we explore data heterogeneity in FL settings, the models, however, are homogenous across clients and the server for a dataset.
\vspace{-2mm}
% data distribution
\paragraph{Dataset Distribution.} Let $\mathcal{D}$ denote the centralized dataset, which serves as the complete collection of data before federated partitioning. The subsets of this dataset $\{\mathcal{D}_{val},\mathcal{D}_{test}\} \subset \mathcal{D}$ are used by the server for validation and testing, respectively of the global model. To simulate the federated setting, $\mathcal{D}_{rem} := \mathcal{D} \setminus \{\mathcal{D}_{val},\mathcal{D}_{test}\}$ is divided into multiple client-specific subsets. If institutional information regarding data collection exists, it is utilized to partition the dataset into $n$ splits, such that $\{\mathcal{D}_1, \mathcal{D}_2, \ldots, \mathcal{D}_n\} \subset \mathcal{D}_{rem}$, constituting a natural partition. In the absence of such information, synthetic partitioning based on metadata and label information is performed to create client datasets under both IID and non-IID settings, with the latter generated using a Dirichlet distribution.
Please refer to~\cref{sec:datasets} to dive deep on how each dataset is partitioned.

% table here
\subsection{FL Algorithms}
Med-MMFL implements and evaluates six representative FL algorithms selected to cover diverse state-of-the-art paradigms with distinct optimization and update mechanisms. Notably, most of these methods were originally designed for unimodal or, at best, bimodal applications. In our work, we extend selected algorithms to support multimodal data, enabling a fair and unified evaluation.
\vspace{-4mm}

\paragraph{FedAvg \cite{McMahan2017Communication}.} FedAvg introduces a FL framework that allows clients to collectively train a global model keeping its local data private. It performs iterative round-based training where at the start of each round $t+1$, each client $k$ receives a copy of the global model $w_t$ and updates it to optimize the local objective \(
\mathcal{L}_{\text{local}} \) for a number of local epochs. At the end of each round, the server receives copies of client models \(w_k, k=1,2..C\) and aggregates them to get new global model $w_\text{t+1}$. Subsequent studies ~\cite{li2020convergencefedavgnoniiddata, 49350hsu, li2022federated} established that this algorithm's convergence rates may degrade under heterogeneity.
\vspace{-4mm}

\paragraph{FedProx \cite{li2020federatedprox}.} FedProx follows the same model-aggregation framework as FedAvg. However, to tackle heterogeneity in federated settings, it improves the local objective by adding a $L_2$ regularization term to each local training loss, which penalizes the deviation of the local models from the last global model. Additionally, it introduces a hyper-parameter $\mu$ to control the effect of the regularization. 
\vspace{-4mm}

\paragraph{SCAFFOLD \cite{karimireddy2021scaffoldstochasticcontrolledaveraging}.} When the distribution of each local dataset differs from the global distribution, there exists a \textit{drift}~\cite{karimireddy2021scaffoldstochasticcontrolledaveraging} in local updates away from the global optimum. This drift induces bias in the aggregated updates, misguiding the global optimization process and consequently slower or unstable convergence of the global model. To address this, SCAFFOLD introduces control variates that estimates the client drift and corrects the gradients in the direction that compensates for the drift. 
\vspace{-4mm}

\paragraph{FedNova \cite{wang2020tacklingobjectiveinconsistencyproblemnova}.}
FedNova improves FedAvg in the aggregation stage by considering that different parties may conduct different numbers of local steps in each communication round and the parties with a larger number of local steps may significantly bias the global updates towards them. Thus, to ensure a more balanced aggregation, FedNova normalizes and scales the local model updates of each client according to the size of their local steps before updating the global model. 
\vspace{-2mm}

\paragraph{MOON \cite{li2021modelcontrastivefederatedlearningMOON} and m-MOON (ours).} 
\label{par:m-Moon}
To mitigate drift under non-IID conditions, MOON uses contrastive learning to align local and global representations. It minimizes the distance between features learned by the local and global models while maximizing the distance from the previous local model. Note that MOON only formulates the algorithm for unimodal data. 
Here, we propose to generalize the model-contrastive learning principle of MOON to more modalities, \emph{i.e.}, multimodal MOON (m-MOON) for evaluation on multimodal datasets. To this end, we perform \textit{modality-wise contrastive alignment} between the local and global representations for each modality $m \in M_C$ (see ~\cref{eq:m-moon}). 
\begin{equation}
\begin{aligned}
    &\mathcal{L}_{m-MOON}  = \sum_{m \in M_C} -\log{\frac{f(\mathbf{z}^m_{loc}, \mathbf{z}^m_{glob})}{f(\mathbf{z}_{loc}^m, \mathbf{z}^m_{glob}) + f(\mathbf{z}_{loc}^m, \mathbf{z}^m_{prev})}} 
\end{aligned}
\label{eq:m-moon}
\end{equation}

where
    $f(\mathbf{z_1},\mathbf{z_2}) = \exp (\textrm{sim}(\mathbf{z_1},\mathbf{z_2})/\tau)$ and $\mathbf{z}^m_{k}$ is a representation of the modality $m$ from the model $k$.
The intuition for this design choice over contrastive learning on a singular fused multi-modal representation stems from the fact that each modality may drift independently during local training due to heterogeneity. To balance the contrastive regularization with the task-specific local objective, m-MOON keeps the hyperparameter $\mu$ from MOON \cite{li2021modelcontrastivefederatedlearningMOON}, that controls the weight of the model-contrastive loss. 
% Furthermore, MOON can be viewed as enforcing representation proximity constraints between local and global models' encoders. When this contrastive objective is applied to the fused multi-modal embedding, it imposes only a single global constraint in a high-dimensional fused latent space. That would give weaker control over the representations geometry of individual modalities compared to m-MOON.
\vspace{-2mm}

\paragraph{CreamFL~\cite{yu2023multimodalfederatedlearningcontrastiveCreamFL} and CreamMFL (ours).}
\label{par:creamfl}
CreamFL enables a server model to be learned from clients with heterogeneous modalities and model architectures. To achieve this, it performs a contrastive representation-level ensemble and a global–local cross-modal aggregation on a small public dataset, allowing the server to fuse client representations while preserving client privacy. The framework additionally introduces inter-modal and intra-modal contrastive regularizers during local training to mitigate local drift that may arise from modality gaps and/or task gaps. However, the formulation of CreamFL is limited to bimodal (image and text) scenarios only. We generalize CreamFL beyond 2 modalities with \textbf{CreamMFL}, by simply extending contrastive regularizer for each modality using pairwise inter- and intra-modal losses (see~\cref{eq:creamMFL}) and global-local contrastive aggregation by using the available modalities.

\begin{subequations}\label{eq:creamMFL}
\begin{equation}
\begin{aligned}
{
\ell^{(r)}_{intra} =
\sum_{m \in M_C} 
- \log 
\frac{f(\mathbf{z}^{(r,m)}, \mathbf{z}_{glob}^{(r,m)})}
     {f(\mathbf{z}^{(r,m)}, \mathbf{z}_{glob}^{(r,m)}) 
    + f(\mathbf{z}^{(r,m)}, \mathbf{z}^{(r,m)}_{prev})}
}
\end{aligned}
\label{eq:creamMFL-intra}
\end{equation}

\begin{equation}
\begin{aligned}
{
\ell^{(r)}_{inter} =
\sum_{\substack{m_1,m_2 \in M_C \\ m_1 \ne m_2}}
- \log 
\frac{
f(\mathbf{z}^{(r,m_1)}, \mathbf{z}_{glob}^{(r,m_2)})
}{
\sum_{\substack{n=1 \\ n \ne r}}^{|\mathcal{P}|}
f(\mathbf{z}^{(r,m_1)}, \mathbf{z}^{(n,m_2)}_{glob})
}
}
\end{aligned}
\label{eq:creamMFL-inter}
\end{equation}
\end{subequations}

where $f(\mathbf{z_1}, \mathbf{z_2}) = \exp(\mathbf{z_1}^\top \cdot\ \mathbf{z_2})$, $\mathbf{z}^{r,m}_{glob}$ is the representation of $r^{th}$ data point of modality $m$ from $glob$ model and $|\mathcal{P}|$ is the public data available to all clients. For fair comparison across consistent client configurations, CreamMFL trains a separate model on the public data as a virtual client which is also used in aggregation. 

\begin{table*}[t]
\centering

 % Text colors in red and blue denote the best performance using \textit{Concatanation-based Fusion} and \textit{Attention-based Fusion}, respectively. $\dagger$ indicates the best performance score of the corresponding dataset.
    % \vspace{ c}
\resizebox{0.9\linewidth}{!}{
\begin{tabular}{cccccccccc}

\toprule
\textbf{Split} &
\textbf{Datasets} &
\textbf{Clients} &
\textbf{Metric $\uparrow$} &
\textbf{FedAvg} &
\textbf{FedProx} &
\textbf{SCAFFOLD} &
\textbf{m-MOON} &
\textbf{FedNova} &
\textbf{CreamMFL} \\
\midrule

% Natural Partition
\multirow{1}{*}{\parbox{1.75cm}{\centering \textbf{Natural}}} &
Fed-BraTS-GLI2024 &
5 &DSC &
81.797 &
81.260 &
79.447 &
\textbf{82.406} &
\underline{82.125} &
78.934 \\
\hline
\midrule
% Synthetic IID
% \rowcolor[HTML]{eff0f1}
\multirow{6}{*}{\parbox{1.75cm}{\centering \textbf{Synthetic\\IID}}} & 

 \multirow{2}{*}{Fed-BraTS-GLI2024} &
3 & DSC & 82.608 & 83.259 & 81.944 & \underline{83.555} & \textbf{84.052} & 76.539 \\
% \rowcolor[HTML]{eff0f1}

& & 5 & DSC & \underline{80.781} & 79.717 & 78.108 & 79.655 & \textbf{82.811} & 77.064 \\
% \rowcolor[HTML]{eff0f1}
% 7 clients omitted for now
% & & 7 & DSC & \textbf{79.905} & 77.620 & 78.099 & 77.940  & - & - \\
\cmidrule{2-10}

& \multirow{2}{*}{Fed-MIMIC-CXR-JPG} &
3 & AUC & 82.996 & \underline{83.245} & \textbf{83.539} & 82.016 & 82.582 & 80.607 \\
& & 5 & AUC & \textbf{80.202} & 78.983 & \underline{79.978} & 77.186  & 79.905 & 77.871 \\
\cmidrule{2-10}

& \multirow{2}{*}{Fed-Symile-MIMIC} &
3 & Acc & \textbf{38.147 }& 35.698 & 26.94\footnotemark[1] & \underline{38.074} & 11.853 & 18.966 \\
% \rowcolor[HTML]{eff0f1}
& & 5 & Acc & \textbf{34.483} & \underline{30.316} & 21.336\footnotemark[1] & 30.244 & 9.6983 & 16.164 \\
\cmidrule{2-10}
% \rowcolor[HTML]{eff0f1}
% 7 clients omitted for now
% & & 7 & Acc & 22.342 & 20.618 & - & \textbf{31.896} & - & - \\
% \rowcolor[HTML]{eff0f1}

& \multirow{1}{*}{Fed-PathVQA} &
3 & F1 & 86.895 & 87.186 & \textbf{87.852} & \underline{87.558} & 84.937 & 84.411 \\
% & & 5 & F1 & \textbf{-} & - & - & -  & - & - \\

% & & 7 & F1 & \textbf{79.905} & 77.620 & 78.099 & 77.940  & - & - \\
\cmidrule{2-10}
& \multirow{1}{*}{Fed-EHRXQA} &
3 & F1 & 51.008 & 51.044 & \textbf{51.552} & \underline{51.4} & 50.706 & 47.524 \\
% & & 5 & F1 & \textbf{-} & - & - & -  & - & - \\

\midrule
\multicolumn{4}{c}{\# times that performs the best}
& 3 & 0 & 3 & 0 & 2 & 0 \\
\hline
\midrule

% Synthetic non-IID alpha 0.8
\multirow{6}{*}{\parbox{1.75cm}{\centering \textbf{Synthetic\\non-IID} \\ ${\alpha = 0.8}$}} 

& \multirow{2}{*}{Fed-BraTS-GLI2024} &
3 & DSC & 81.536 & 80.518 & \textbf{83.941} & \underline{83.260} & 83.164 & 80.592 \\
& & 5 & DSC & 80.292 & \underline{80.466} & 79.381 & 80.432 & \textbf{82.325} & 80.069 \\
% 7 clients omitted for now
% & & 7 & DSC & - & - & - & - & - & - \\
\cmidrule{2-10}

& \multirow{2}{*}{Fed-MIMIC-CXR-JPG} &
3 & AUC & \underline{83.763} & 83.063 & 83.553 & 82.745 & 82.534 & \textbf{84.777} \\
& & 5 & AUC & \underline{81.649} & \textbf{81.741} & 79.717 & 79.714  & 79.608 & 71.227 \\
\cmidrule{2-10}
& \multirow{2}{*}{Fed-Symile-MIMIC} &
3 & Acc & \underline{32.902} & \textbf{35.345} & 15.086\footnotemark[1] & 24.856 & 11.207 & 17.026 \\
& & 5 & Acc & \textbf{31.824} & \underline{30.244} & 19.181\footnotemark[1] & 29.597 & 10.991 & 14.224 \\
% & & 7 & Acc & \textbf{33.261} & 28.951 & - & 30.244 & - & - \\
\cmidrule{2-10}

& \multirow{1}{*}{Fed-PathVQA} &
3 & F1 & \textbf{87.022} & \underline{86.99} & 84.978 & 85.863 & 85.194 & 85.764 \\
% & & 5 & F1 & - & - & - & - & - & - \\
\cmidrule{2-10}

& \multirow{1}{*}{Fed-EHRXQA} &
3 & F1 & 50.542 & \textbf{51.141} & 50.793 & \underline{50.993} & 49.802 & 46.038 \\
% & & 5 & F1 & - & - & - & - & - & - \\

\midrule

\multicolumn{4}{c}{\# times that performs the best}
& 2 & 3 & 1 & 0 & 1 & 1 \\
\hline
\midrule

% Synthetic non-IID alpha 0.2
\multirow{6}{*}{\parbox{1.75cm}{\centering \textbf{Synthetic\\non-IID} \\ ${\alpha = 0.2}$}} 

& \multirow{2}{*}{Fed-BraTS-GLI2024} &
3 & DSC & 81.779 & \textbf{83.146} & 81.658 & 83.011 & \underline{82.896} & 81.766 \\
& & 5 & DSC & 81.171 & 81.121 & 80.628 & \underline{81.250} & \textbf{83.579} & 74.817 \\
% & & 7 & DSC & - & - & - & - & - & - \\
\cmidrule{2-10}

& \multirow{2}{*}{Fed-MIMIC-CXR-JPG} &
3 & AUC & \textbf{84.108} & 82.731 & \underline{82.884} & 82.157 & 82.447 & 78.827 \\
& & 5 & AUC & 82.745 & \textbf{83.793} & 77.252 & 81.755  & 81.098 & 73.969 \\
\cmidrule{2-10}

& \multirow{2}{*}{Fed-Symile-MIMIC} &
3 & Acc & \underline{33.333} & 28.736 & 11.638\footnotemark[1] & \textbf{34.195} & 11.207 & 13.362 \\
& & 5 & Acc & \underline{31.394} & \textbf{34.698} & 17.886\footnotemark[1] & 22.773 & 9.2672 & 18.75 \\
% & & 7 & Acc & 29.238 & \textbf{29.598} & - & 9.555 & - & - \\
\cmidrule{2-10}

& \multirow{1}{*}{Fed-PathVQA} &
3 & F1 & 87.155 & \underline{87.176} & \textbf{87.428} & 85.813 & 85.123 & 86.224 \\
% & & 5 & F1 & - & - & - & - & - & - \\
\cmidrule{2-10}

& \multirow{1}{*}{Fed-EHRXQA} &
3 & F1 & \underline{51.445} & \textbf{51.549} & 51.170 & 50.993 & 49.751 & 47.990 \\
% & & 5 & F1 & - & - & - & - & - & - \\

\midrule

\multicolumn{4}{c}{\# times that performs the best}
& 1 & 4 & 1 & 1 & 1 & 0 \\
\hline

\bottomrule
\end{tabular}}
\captionsetup{width=0.85\linewidth}
\caption{MMFL benchmark results reported for 6 FL algorithms across different multimodal medical datasets and partitions. For a setting, \textbf{bold} highlights the best performance, whereas \underline{underlined} highlights the second-best.}
\vspace{-3mm}
\label{table:baseline_results}
\end{table*}

%% file: sec/5_experiments.tex
\section{Experiments}
\label{sec:experiments}

\subsection{Experimental Setup}
To evaluate the performance of FL algorithms across medical datasets, we conduct extensive experiments across different partitioning scenarios on five federated multimodal medical datasets, as described in~\cref{sec:datasets}. 
\vspace{-2mm}
\paragraph{Implementation details.} 
All datasets are first divided into training, validation, and test subsets prior to federated partitioning, ensuring consistent evaluation data across centralized and FL experiments. 
Focusing on cross-silo settings, all experiments are performed under full client participation. The Adam optimizer~\cite{kingma2017adammethodstochasticoptimization} is employed for local optimization, except in the case of FedNova~\cite{wang2020tacklingobjectiveinconsistencyproblemnova}. Since FedNova does not define the normalization vector $\|a\|_1$~\cite{wang2020tacklingobjectiveinconsistencyproblemnova} when using Adam, we instead use SGD with momentum and tune the learning rate independently. The complete set of hyperparameter values is included in the supplementary material.
All algorithms are executed with identical training schedules, including the same number of communication rounds and local epochs. For better reliability, results are averaged over three independent runs with different random seeds. 
All experiments are performed on NVIDIA A100-PCIE-40GB GPUs.

\subsection{Centralized Baseline}
\label{par:central}
\begin{table}[h]
  \centering
  \resizebox{\columnwidth}{!}{
    \begin{tabular}{@{}|lcc|c|c|@{}}
      \hline
      \textbf{Dataset} & \textbf{Metric $\uparrow$} & \textbf{\# Epochs} & \textbf{Centralized} & \textbf{FL Evaluation (min, max)}\\
      \hline
      \hline
      BraTS-GLI2024 \cite{deverdier20242024braintumorsegmentation} & DSC & 150 &84.900 & (74.817, 84.052) \\
     MIMIC-CXR-JPG \cite{johnson2019mimiccxrjpglargepubliclyavailable} & AUC & 40 & 92.380 & (71.227, 84.777) \\
      Symile-MIMIC \cite{saporta2025symilemimic} & Acc & 50 & 41.370 & (9.2672, 38.147) \\

      PathVQA \cite{he2020pathvqa30000questionsmedical} & F1 & 60 & 87.124 & (84.411, 87.852) \\
      EHRXQA \cite{bae2023ehrxqamultimodalquestionanswering} & F1 & 30 & 52.385 & (46.038, 51.552) \\
    \hline
    \end{tabular}
    }
  \captionsetup{width=0.9\columnwidth}
  \caption{Centralized baseline performance obtained by aggregating all client data for reference comparison with federated settings.}
  \label{tab:central_baseline}
  \vspace{-1mm}
\end{table}

 ~\cref{tab:central_baseline} reports the centralized baseline obtained by training on the data aggregated from the clients. While this setup is typically unrealistic in federated contexts due to privacy and governance constraints~\cite{Xu2021Federated}, it offers a useful comparison to evaluate how data decentralization and non-IID distributions impact model performance. Across most datasets, the difference between centralized and federated training is minimal, typically less than 1--1.5\%. An exception is MIMIC-CXR-JPG, where centralized training achieves nearly 7\% higher accuracy than the best-performing federated algorithm. Interestingly, federated algorithms outperform the centralized baseline in certain scenarios, consistent with trends observed in prior studies~\cite{zheng2025fedvlmbenchbenchmarkingfederatedfinetuning, Yeom2024, Alasmari2025}. This behavior may be attributed to the heterogeneous nature of federated updates. Unlike centralized training, which optimizes the model over the entire global dataset, federated training aggregates updates computed on diverse local data distributions. In non-IID settings, each client’s local optimization effectively constitutes a distinct stochastic process, rather than identical copies of the same optimization as assumed under the IID premise. This diversity introduces implicit regularization through stochasticity in the global update, guiding the model toward flatter regions of the loss landscape and improving generalization~\cite{keskar2017largebatchtrainingdeeplearning, Li_2020}.

\renewcommand{\thefootnote}{\fnsymbol{footnote}}
\footnotetext[1]{Gradient scaling was applied to control the gradient correction in  SCAFFOLD for Fed-Symile-MIMIC.} 
 
% Required to mention the above, as the above is clearly seen in this table
\subsection{Main Results on Med-MMFL}
\begin{figure}[t]
    \centering
    \includegraphics[width=0.90\linewidth]{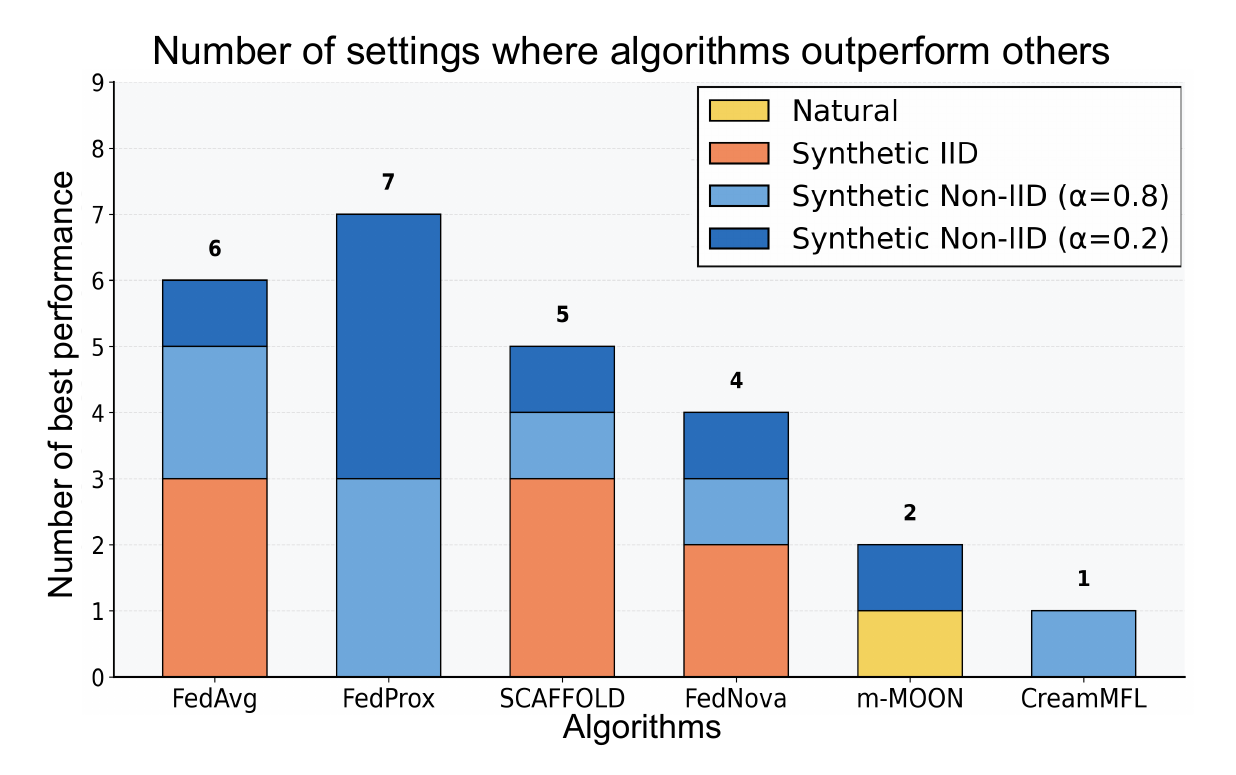}
    \caption{Number of settings in which each algorithm outperforms the others across our Med-MMFL benchmark. The stacked bars are color-coded by data partition type.}
    \label{fig:algorithm_outperform_count} 
    \vspace{-6mm}
\end{figure}

The performance of six FL algorithms, reported across varying partitions and datasets are shown in ~\cref{table:baseline_results}. Next, we present an in-depth analysis and insights of these results from different perspectives.
% \vspace{-4mm}
\paragraph{Comparison among algorithms.}
Overall, no single algorithm consistently outperforms others across all experimental configurations. In~\cref{table:baseline_results}, we noticed that FedProx, FedAvg, and SCAFFOLD consistently rank among the top-performing algorithms across diverse combinations of datasets, data splits, and class counts, albeit in different contexts. FedProx achieves the highest performance in 7 out of 16 settings, particularly excelling in non-IID scenarios as highlighted by ~\cref{fig:algorithm_outperform_count}. In contrast, FedAvg and SCAFFOLD demonstrate their strongest performance primarily in IID settings. These results suggest that in real-world settings, where datasets are inherently heterogenous, FedProx is likely to offer more robust performance. This also aligns with prior works~\cite{49350hsu, li2020convergencefedavgnoniiddata} that establish the degradation in performance of FedAvg with increase in heterogeneity. Interestingly, although FedNova does not rank first or second in most benchmarks, it performs exceptionally well on the Fed-BraTS-GLI2024 dataset, achieving the best results in 4 settings and the second-best in 2 settings, totaling top performances in 6 out of 7 configurations. 

\paragraph{Effect of data partitioning strategy on performance.}
The effect of data partitioning strategy on model performance remains generally consistent across most datasets, with notable exceptions. For Fed-BraTS-GLI2024 and Fed-MIMIC-CXR-JPG, performance remains largely consistent across splits, except for CreamMFL, which exhibits a variation of up to 5–6\% in the 3-client setting. In contrast, Fed-Symile-MIMIC exhibits a significant drop in performance when moving from IID to non-IID configurations across all algorithms. This decline is likely due to the nature of contrastive learning, which relies on diverse batch samples to effectively push apart embeddings of different instances while pulling together the embeddings of the same instance. Under IID splits, each client receives a representative mix of data, allowing the model to learn discriminative features across all types of examples. However, in non-IID splits, some data groups may contain only a few samples. In such cases, the model has limited examples to learn from, making it difficult to distinguish instances within that group and reducing overall embedding quality. Conversely, Fed-PathVQA and Fed-EHRXQA demonstrate robustness to data heterogeneity, maintaining comparable performance across different splits.
\vspace{-3mm}
\paragraph{Best-performing algorithms across datasets.}
The top-performing algorithms vary across datasets, reflecting that each method tends to excel under specific conditions. On Fed-BraTS-GLI2024, FedNova performs best in 4 out of 7 settings. For Fed-MIMIC-CXR-JPG, FedAvg and FedProx lead in 2 settings each, together covering 4 of 6 configurations. In 
Fed-Symile-MIMIC, FedAvg dominates 3 settings and FedProx 2, while SCAFFOLD and FedNova perform notably worse. As recent work suggests that Self-supervised Learning (SSL) gradients possess predictive capabilities and encode complementary information beyond model 
embeddings~\cite{simoncini2024traingainselfsupervisedgradients}, hence, this degradation may stem from instability introduced by frequent gradient corrections or cumulative gradient-based normalization. On Fed-PathVQA, SCAFFOLD achieves the best results in 2 of 3 settings, while 
on Fed-EHRXQA, FedProx leads in 2 out of 3 cases, highlighting the effectiveness of regularization via weight and gradient control in VQA tasks. Overall, while each dataset favors different algorithms, FedAvg and FedProx emerge as the most consistently strong performers across this experimental setup, ranking highest in 3 out of 5 datasets.

%% file: sec/6_conclusion.tex
\section{Conclusion}
\label{sec:conclusion}
We present Med-MMFL, the first comprehensive benchmark for multimodal federated learning (MMFL) in healthcare. Our benchmark integrates 6 state-of-the art FL algorithms, 4 types of clinical tasks, 3 data partitioning strategies, and 10 unique medical modalities across 5 datasets. Although Med-MMFL covers a broad spectrum of FL settings, some relevant algorithms remain beyond its current scope and are left for future exploration. Med-MMFL focuses on the federated aspects of multimodal learning, emphasizing the aggregation of client updates rather than the design or evaluation of fusion strategies. A systematic study of fusion methods is therefore beyond the scope of this work. Nonetheless, by publicly releasing the benchmark, Med-MMFL aims to foster transparency, reproducibility, and consistent comparisons for future research. We anticipate that this benchmark will accelerate the development of robust, clinically relevant multimodal FL solutions and serve as a valuable resource for the medical AI community.

%% file: sec/X_suppl.tex
\clearpage
\setcounter{page}{1}
\maketitlesupplementary
In this supplementary material, we first provide additional information about each dataset individually, including details on the corresponding data partitioning strategy, in ~\cref{sec:supp_datasets}. 
% \cref{sec:supp_FL} offers an expanded overview of the federated learning algorithms evaluated in our study. 
In ~\cref{sec:supp_experiments}, we describe further details regarding the experimental setup, including hyperparameter configurations. ~\cref{sec:supp_results} presents additional results, covering both quantitative and qualitative components. 
% Finally, ~\cref{sec:supp_convergence} includes convergence plots for the different algorithms considered in this work.

\section{Datasets and Partitioning Strategy}
\label{sec:supp_datasets}

\paragraph{Fed-BraTS-GLI2024.} 
We evaluate the methods on the multimodal brain tumor segmentation task in Fed-BraTS-GLI2024. The objective is to segment three tumor sub-regions: \textbf{whole tumor}, \textbf{tumor core}, and \textbf{enhancing tumor} from multimodal MRI scans. The whole tumor comprises all three constituent sub-regions: the necrotic and non-enhancing tumor core (NCR/NET), the peritumoral edema (ED), and the GD-enhancing tumor (ET). The tumor core includes NCR/NET and ET, whereas the enhancing tumor corresponds exclusively to ET.

\begin{figure}[ht]
    \centering
    \includegraphics[width=\linewidth]{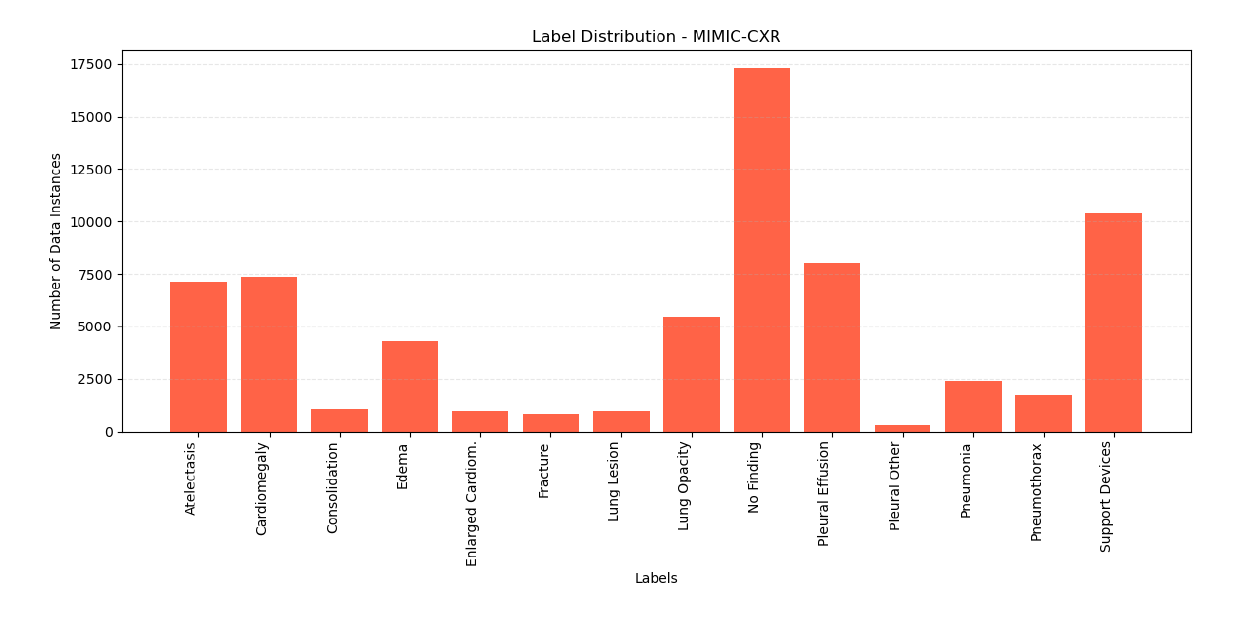}
    \caption{The distribution of labels in MIMIC-CXR-JPG~\cite{johnson2019mimiccxrjpglargepubliclyavailable}}
    \label{fig:mimiclabels}
\end{figure}

\paragraph{Fed-MIMIC-CXR-JPG.}
Fed-MIMIC-CXR-JPG is evaluated on a multi-label classification task. ~\cref{fig:mimiclabels} illustrates the distribution of the 14 pathology labels in the dataset, showing the frequency of instances annotated with: Atelectasis, Cardiomegaly, Consolidation, Edema, Enlarged Cardiomediastinum, Fracture, Lung Lesion, Lung Opacity, No Finding, Pleural Effusion, Pleural Other, Pneumonia, Pneumothorax, and Support Devices.

\paragraph{Fed-EHRXQA.}
As described in~\cref{sec:datasets}, the original EHRXQA dataset~\cite{bae2023ehrxqamultimodalquestionanswering} contains multi-patient SQL-style queries that retrieve information across the entire EHR database. Our goal, however, is to focus the evaluation on tasks that naturally reflect clinical reasoning, where a model interprets multimodal information for a single patient. Population-level SQL retrieval such as counting, listing, or filtering patients, does not align with this objective. To align the dataset with this objective, we removed templates requiring database-wide aggregation (e.g., listing or counting all patients matching some condition). Representative examples of discarded templates are shown in ~\cref{tab:ehrxqa_removed}, where both the template and an actual sample question are included.

\begin{table*}[ht]
\centering
\small

\begin{tabular}{p{0.46\linewidth} p{0.46\linewidth}}
\toprule
\textbf{Template} & \textbf{Sample Question} \\
\midrule
\texttt{list the ids of patients who had any chest x-ray study indicating \${attribute} [time\_filter\_global1].}
&
\texttt{identify the patients by their ids who had chest x-ray studies indicating low lung volumes since 03/2103.}
\\[1em]
\hline
\texttt{count the number of patients aged [age\_group] who had a chest x-ray study during hospital visit indicating any \${category} in the \${object} [time\_filter\_global1].}
&
\texttt{how many patients in the 20s age group had chest x-ray findings of any anatomical findings in the right lung until 05/2101?}
\\
\bottomrule
\end{tabular}
\caption{Examples of removed EHRXQA question templates that require multi-patient SQL-style retrieval.}
\label{tab:ehrxqa_removed}
\end{table*}

To ensure that the task aligns with realistic, single-patient clinical VQA, we retain only those templates explicitly anchored to a \texttt{patient\_id} or \texttt{study\_id}. These can be answered using a single patient's EHR and corresponding chest X-ray. Examples of retained templates and their sample questions are shown in ~\cref{tab:ehrxqa_retained}. This allows us to preserve rich, per-patient EHR data and support meaningful multimodal reasoning between EHRs and CXRs. The retained instances naturally constrain the task to single-patient, multimodal clinical VQA (see~\cref{fig:ehrxqa_qual}), eliminating the need for database-wide SQL retrieval. For a full description of the original EHRXQA schema and query design, we refer readers to the dataset's accompanying publication~\cite{bae2023ehrxqamultimodalquestionanswering}.

\begin{figure*}[!ht]
    \centering

    \begin{subfigure}{0.33\linewidth}
        \centering
        \includegraphics[width=\linewidth]{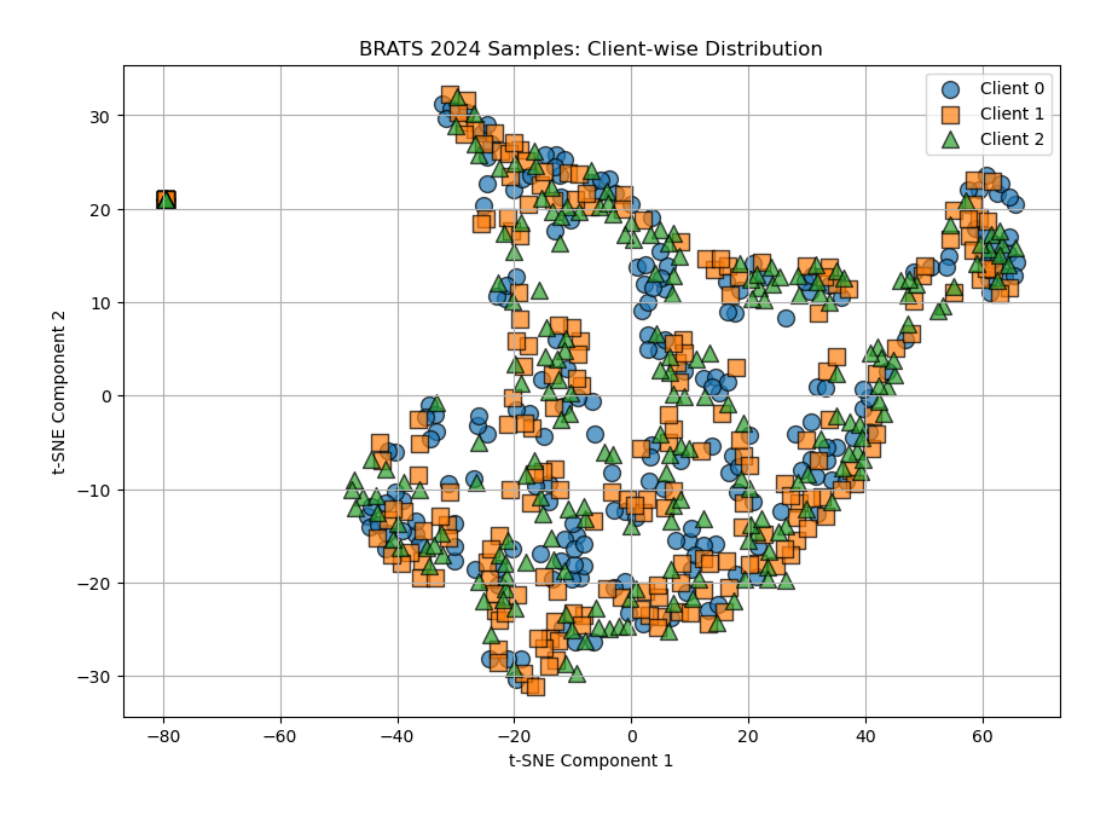}
        \captionsetup{width=0.9\linewidth}
        \caption{Synthetic IID partition: Class proportion vectors are uniformly distributed across clients.}
        \label{fig:brats_homoc3}
    \end{subfigure}
    \hfill
    \begin{subfigure}{0.33\linewidth}
        \centering
        \includegraphics[width=\linewidth]{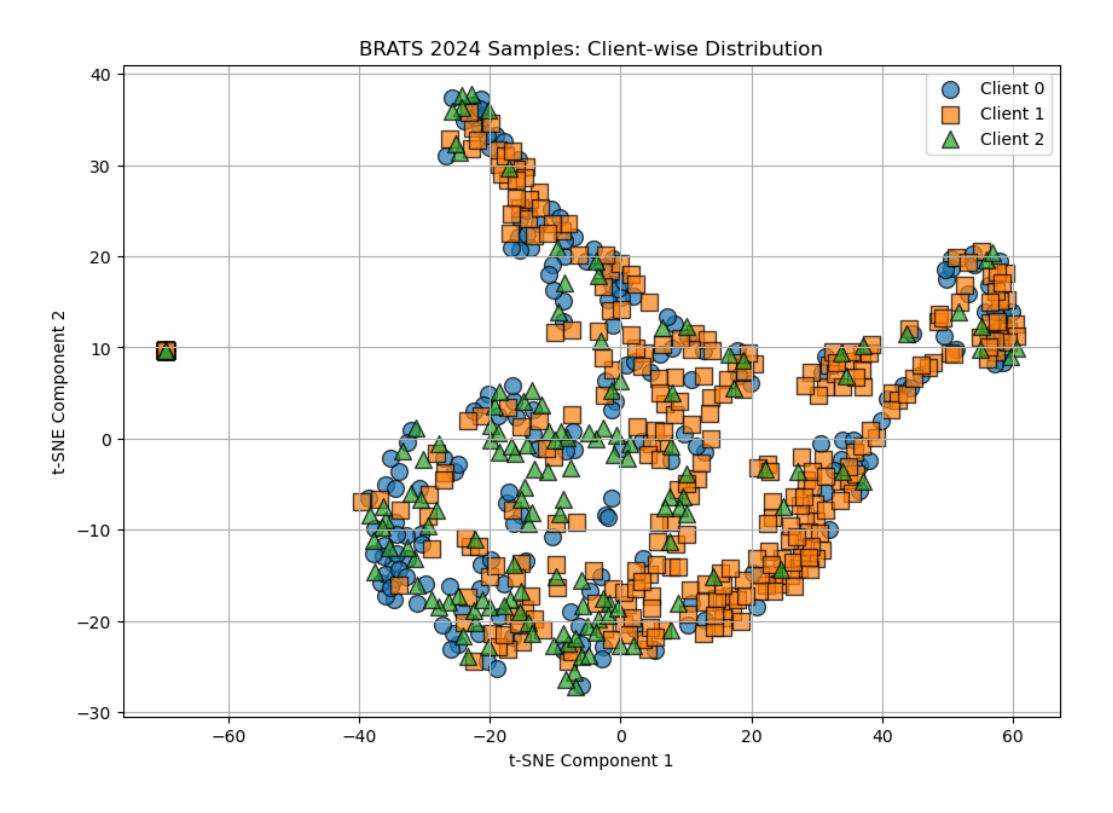}
        \captionsetup{width=0.9\linewidth}
        \caption{Synthetic non-IID partition ($\alpha = 0.8$): moderate heterogeneity yields partially separated clusters, indicating uneven pseudo-class exposure across clients.}

        \label{fig:brats_c3a08}
    \end{subfigure}
    \hfill
    \begin{subfigure}{0.33\linewidth}
        \centering
        \includegraphics[width=\linewidth]{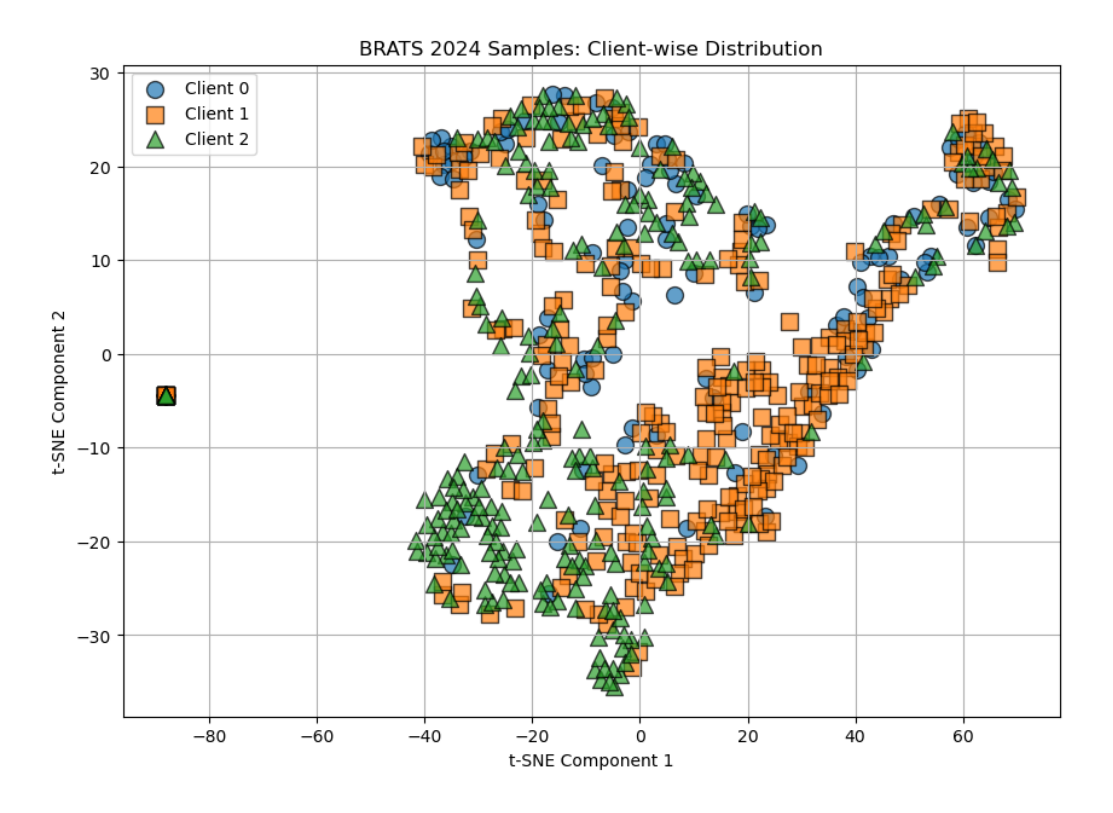}
        \captionsetup{width=0.\linewidth}
        \caption{Synthetic non-IID partition ($\alpha = 0.2$): strong heterogeneity produces distinct, well-separated client distributions, consistent with highly skewed pseudo-class allocation.}

        \label{fig:brats_c3a02}
    \end{subfigure}
    \captionsetup{width=0.95\linewidth}
    \caption{t-SNE visualization of class-proportion vectors across clients for the Fed-BraTS-GLI2024 dataset under three partitioning strategies: synthetic IID, synthetic non-IID ($\alpha=0.8$), and synthetic non-IID ($\alpha=0.2$), each with 3 clients. Class-proportion vectors are derived from normalized voxel counts per class within each segmentation mask. As the Dirichlet concentration parameter $\alpha$ decreases, client distributions become increasingly separated, illustrating the controlled progression from balanced to highly skewed label distributions.}
    \label{fig:brats_dist}
\end{figure*}

\begin{figure*}[t]
    \centering

    \begin{subfigure}{0.48\linewidth}
        \centering
        \includegraphics[width=\linewidth]{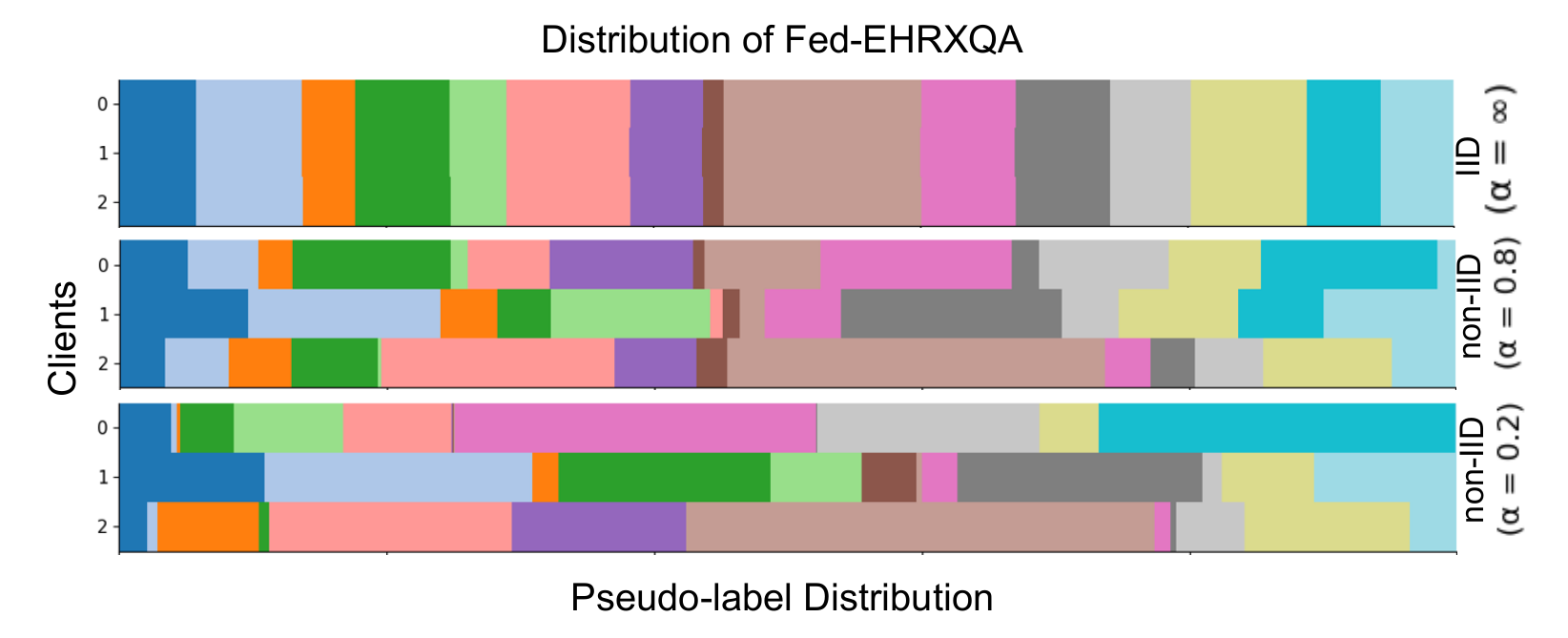}
        \captionsetup{width=0.9\linewidth}
        \caption{Fed-EHRXQA pseudo-label distribution. IID splits yield similar client-level frequencies, while non-IID splits result in heterogeneous distributions.}
        \label{fig:label_distbn_ehrxqa}
    \end{subfigure}
    \hfill
    \begin{subfigure}{0.48\linewidth}
        \centering
        \includegraphics[width=\linewidth]{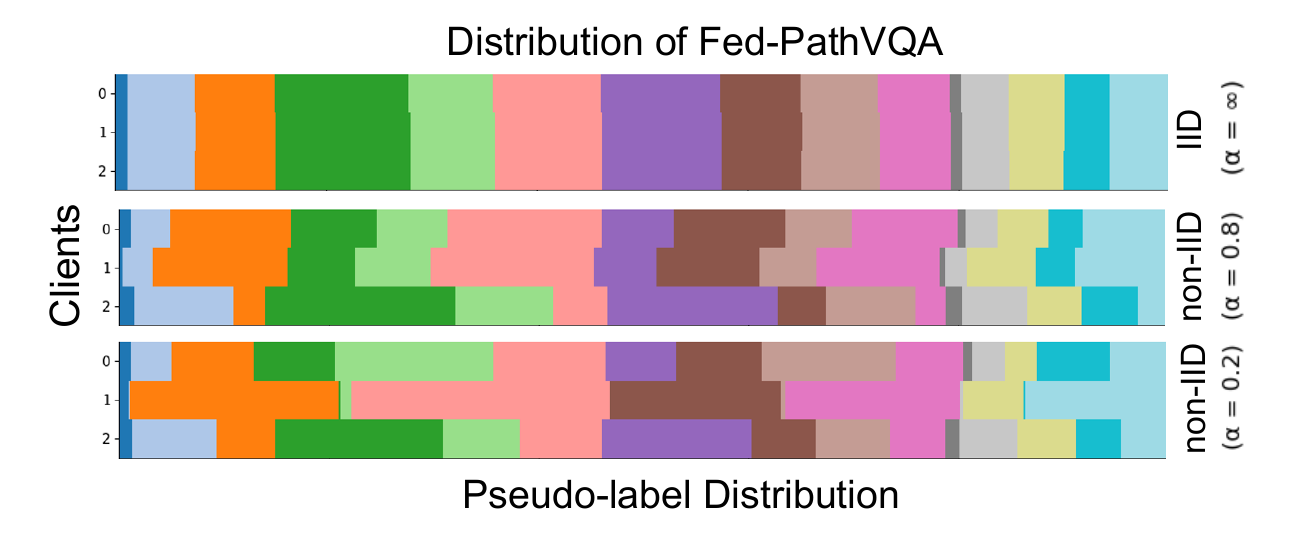}
        \captionsetup{width=0.9\linewidth}
        \caption{Fed-PathVQA pseudo-label distribution under the same synthetic partitioning strategy.}
        \label{fig:label_distbn_pathvqa}
    \end{subfigure}

    \caption{Client-level pseudo-label distributions for Fed-EHRXQA and Fed-PathVQA obtained using our synthetic federated partitioning strategy.}
    \label{fig:label_distbn}
\end{figure*}

\paragraph{Synthetic Partitioning.}For our experiments with synthetic partitions, we follow the Dirichlet-based partitioning strategy described in~\cref{sec:datasets}.  When $\alpha$ is very high (approaching $\infty$), the sampled probabilities $p_k$ are nearly uniform across clients, resulting in roughly equal proportions of each class being assigned to every client. This produces nearly homogeneous splits, which we refer to as \textit{synthetic IID} partitions. Conversely, as $\alpha$ decreases, the sampling becomes increasingly skewed, with certain clients receiving disproportionately more instances of some pseudo-classes than others. This leads to heterogeneous, non-IID distributions across clients, which we denote as \textit{synthetic non-IID}. In our experiments, $\alpha = 0.2$ represents a more heterogeneous partition than $\alpha = 0.8$. 
For a qualitative understanding of the distributional differences introduced by our synthetic Dirichlet partitions in Fed-BraTS-GLI2024, we reduce the class-proportion vectors to two dimensions using t-SNE and visualize them in~\cref{fig:brats_dist}. The plots highlight the transition from balanced to highly skewed client distributions as the Dirichlet parameter decreases. Similarly, client-level distributions of Fed-EHRXQA and Fed-PathVQA resulting from our federated partitioning strategy are illustrated in ~\cref{fig:label_distbn}.

\begin{table*}[ht]
\centering
\small

\begin{tabular}{p{0.46\linewidth} p{0.46\linewidth}}
\toprule
\textbf{Template} & \textbf{Sample Question} \\
\midrule

\texttt{has\_verb patient \{patient\_id\} been prescribed with \{drug\_name\} [time\_filter\_global1] and also had a chest x-ray study indicating \${attribute} within the same period?}
&
\texttt{was patient 11887414 prescribed milk of magnesia since 91 months ago and had a chest x-ray showing a chest port within the same timeframe?}
\\[1em]
\hline
\texttt{given the \{study\_id1\} study, list all anatomical locations related to any \${category} that are \${comparison} compared to the previous study?}
&
\texttt{enumerate all anatomical locations related to any diseases still present in the 50978999 study relative to the previous one.}
% \\[1em]
% \hline
% \texttt{has\_verb patient \{patient\_id\} received a \{procedure\_name\} procedure [time\_filter\_global1] and also had a chest x-ray study indicating any abnormality within the same period?}
% &
% \texttt{did patient 13385073 undergo a single-vessel procedure in 2105 and have a chest x-ray revealing any abnormalities?}
\\
\bottomrule
\end{tabular}
\caption{Examples of retained EHRXQA templates suitable for patient-level clinical VQA.}
\label{tab:ehrxqa_retained}
\end{table*}

\section{Experimental Settings}
\label{sec:supp_experiments}

\begin{table*}[ht]
    \centering
    \resizebox{0.85\linewidth}{!}{%
    \begin{tabular}{|l|c|c|c|c|c|c|c|c|}
        \hline
        \textbf{Dataset} & 
        \makecell{\textbf{Local}\\\textbf{epochs}} &
        \makecell{\textbf{Communication}\\\textbf{rounds}} &
        \makecell{\textbf{Learning}\\\textbf{rate}} &
        \textbf{$\mu_\text{fedprox}$} &
        \textbf{$\mu_\text{moon}$} &
        \textbf{$\tau_\text{moon}$} &
        \textbf{$\gamma_\text{creamfl}$} &
        \textbf{$\alpha_\text{creamfl}$} \\
        \hline
        Fed-BraTS-GLI2024 & 3 & 50 & 0.0002 & 0.1 & 0.1 & 0.5 & 0.002 & 0.03 \\
        Fed-MIMIC-CXR-JPG & 3 & 30 & 0.0001 & 0.1 & 0.1 & 0.5 & 0.002 & 0.03 \\
        Fed-Symile-MIMIC  & 3 & 30 & 0.001  & 0.1 & 0.1 & 0.5 & 0.002 & 0.03 \\
        Fed-PathVQA        & 3 & 20 & 0.00001 & 0.1 & 10 & 0.5 & 0.002 & 0.03 \\
        Fed-EHRXQA        & 5 & 30 & 0.00001 & 0.1 & 0.1 & 0.5 & 0.002 & 0.03 \\        
        \hline
    \end{tabular}%
    }
    \captionsetup{width=0.85\linewidth}
    \caption{Hyperparameters used for FL evaluations.}
    \label{tab:hyperparams}
\end{table*}

For each dataset, we first implement the centralized baseline. To achieve this, we either tune the hyperparameters in the centralized setting or adopt the hyperparameters used in the corresponding original reference implementations (see~\cref{sec:datasets}). Except for FedNova~\cite{wang2020tacklingobjectiveinconsistencyproblemnova}, where the local optimizer differs, we use a single shared set of hyperparameters for each dataset across all FL algorithms. In addition, we tune the algorithm-specific hyperparameters for each FL method using one data partition per dataset and then apply the same tuned values to all remaining partitions. This procedure is intended to ensure consistency and allow us to reliably assess how variations in data partitions influence the performance of FL algorithms under different multimodal medical settings. The complete hyperparameter configuration is provided in~\cref{tab:hyperparams}. For FedNova, the learning rate is tuned separately over the set \{0.1, 0.01, 0.001\}. In contrast, the search spaces for both $\mu_{\text{moon}}$ and $\mu_{\text{fedprox}}$ are \{0.1, 1, 10\}. We directly set $\tau_{\text{moon}} = 0.5$ without additional tuning, following the original work~\cite{li2021modelcontrastivefederatedlearningMOON}, which reported relatively low sensitivity to this hyperparameter.

\begin{table*}[!ht]
    \centering
    \resizebox{0.9\linewidth}{!}{
    \begin{tabular}{ccccccccc}
    \toprule
    
\textbf{Split} &
\textbf{Clients} &
\textbf{Sub-region} &
\textbf{FedAvg} &
\textbf{FedProx} &
\textbf{SCAFFOLD} &
\textbf{m-MOON} &
\textbf{FedNova} &
\textbf{CreamMFL} \\

    \midrule

\multirow{3}{*}{Natural} &
\multirow{3}{*}{5} &
Whole & 81.902 & 82.000 & 78.904 & \textbf{83.224} & 82.664 & 80.958 \\
& & Core & 81.606 & 80.730 & 79.556 & \textbf{81.974} & 81.704 & 77.928 \\
& & Enhancing & 81.883 & 81.049 & 79.880 & \textbf{82.020} & 82.007 & 77.919 \\

    \hline
    \midrule
\multirow{6}{*}{\makecell{Synthetic\\IID}} & 
\multirow{3}{*}{3} &
Whole & 79.778 & 80.972 & 81.062 & 83.595 & \textbf{84.210} & 76.293 \\
& & Core & 83.628 & \textbf{84.081} & 82.146 & 83.215 & 81.865 & 76.435 \\
& & Enhancing & 84.419 & \textbf{84.722} & 82.623 & 83.853 & 82.359 & 76.889 \\

    \cmidrule{2-9}
    
& \multirow{3}{*}{5} & 
Whole & 80.392 & 79.878 & 78.502 & 79.635 & \textbf{84.210} & 77.393 \\
& & Core & 80.415 & 79.399 & 77.731 & 79.471 & \textbf{81.865} & 76.723 \\
& & Enhancing & 81.537 & 79.873 & 78.092 & 79.858 &\textbf{ 82.359 }& 77.075 \\

    \hline
    \midrule
\multirow{6}{*}{\makecell{Synthetic\\non-IID\\$\alpha = 0.8$}} & 
\multirow{3}{*}{3} &
Whole & 81.135 & 79.935 & 83.686 & \textbf{84.022}  & 83.982 & 78.346 \\
& & Core & 81.487 & 80.591 & \textbf{83.820} & 82.707  & 82.524 & 81.346 \\
& & Enhancing & 81.985 & 81.027 & \textbf{84.318} & 83.052  & 82.988 & 82.085 \\

    \cmidrule{2-9}
    
& \multirow{3}{*}{5} & 
Whole & 79.459 & 80.449 & 77.275 & 81.407 & 84.320 & \textbf{84.743} \\
& & Core & 80.531 & 80.250 & 80.194 & 79.725 & \textbf{81.169 }& 77.911 \\
& & Enhancing & 80.886 & 80.699 & 80.674 & 80.164 & \textbf{81.487} & 78.153 \\

    \hline
    \midrule
\multirow{6}{*}{\makecell{Synthetic\\non-IID\\$\alpha = 0.2$}} & 
\multirow{3}{*}{3} &
Whole & 81.258 & 81.678 & 82.668 & 83.011 & \textbf{83.225} & 79.716 \\
& & Core & 81.965 & \textbf{83.738} & 81.066 & 82.923 & 82.671 & 82.689 \\
& & Enhancing & 82.116 & \textbf{84.023} & 81.241 & 83.362 & 82.850 & 82.894 \\

    \cmidrule{2-9}
    
& \multirow{3}{*}{5} & 
Whole & 80.006 & 80.834 & 80.419 & 80.898 & \textbf{84.499} & 74.015 \\
& & Core & 81.333 & 81.034 & 80.501 & 81.207 & \textbf{82.997} & 74.875 \\
& & Enhancing & 82.172 & 81.494 & 80.964 & 81.645 & \textbf{83.240} & 75.562 \\

    \hline
    \midrule
    
    \end{tabular}%
}
    \captionsetup{width=0.9\linewidth}
    \caption{Supplementary per–tumor-subregion Dice scores for the Fed-BraTS-GLI2024 dataset. These metrics complement the average Dice scores reported in~\cref{sec:experiments} by detailing the whole tumor, tumor core, and enhancing tumor regions.}
    \label{tab:subregions}    

\end{table*}

\section{Additional Results}
\label{sec:supp_results}
This section provides supplementary results corresponding to the experiments reported in~\cref{sec:experiments}.

\subsection{Quantitative Results}

\paragraph{Fed-BraTS-GLI2024.} Following the definitions of the whole tumor, tumor core, and enhancing tumor regions provided in~\cref{sec:supp_datasets}, we present in~\cref{tab:subregions}, the per-subregion Dice scores for all six FL algorithms. These results correspond to the same experimental settings reported in the main paper. Notable observations from the per–tumor-subregion Dice scores include the following: in the synthetic IID (3-client) setting, FedProx reports higher Dice scores for the tumor core and enhancing regions, while FedNova achieves the highest overall average. Similarly, in the synthetic non-IID $\alpha = 0.2$ case, FedProx achieves the highest overall average but has a lower whole-tumor score than four other algorithms, with its average largely influenced by stronger tumor-core and enhancing-region performance again. Under the synthetic non-IID configuration with $\alpha = 0.8$, m-MOON attains a higher whole-tumor score than SCAFFOLD, despite SCAFFOLD yielding the better overall average. Although CreamMFL does not outperform any method in terms of overall Dice averages, it records a higher whole-tumor score than FedNova in the synthetic non-IID $\alpha = 0.8$ setting. These patterns indicate that overall Dice averages can obscure sub-region–specific behavior, and that algorithmic performance may vary substantially across tumor sub-regions.

\subsection{Qualitative Results}
\begin{figure*}[t]
    \centering
    \includegraphics[width=\linewidth]{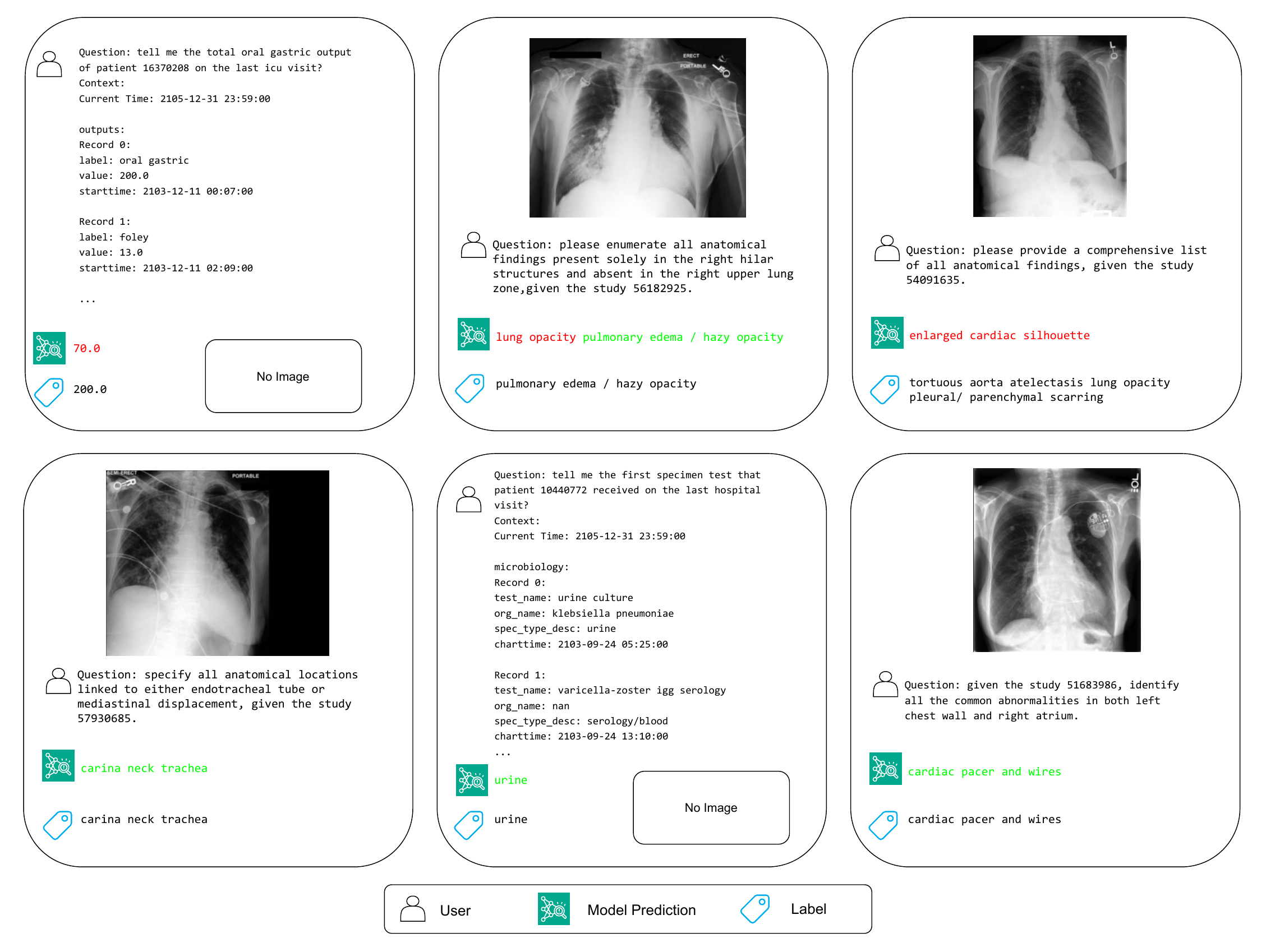}

    \caption{
Qualitative examples from the centralized EHRXQA baseline on patient-level open-ended VQA task.}
    \label{fig:ehrxqa_qual}
\end{figure*}

\begin{figure*}
    \centering
    \includegraphics[width=\linewidth]{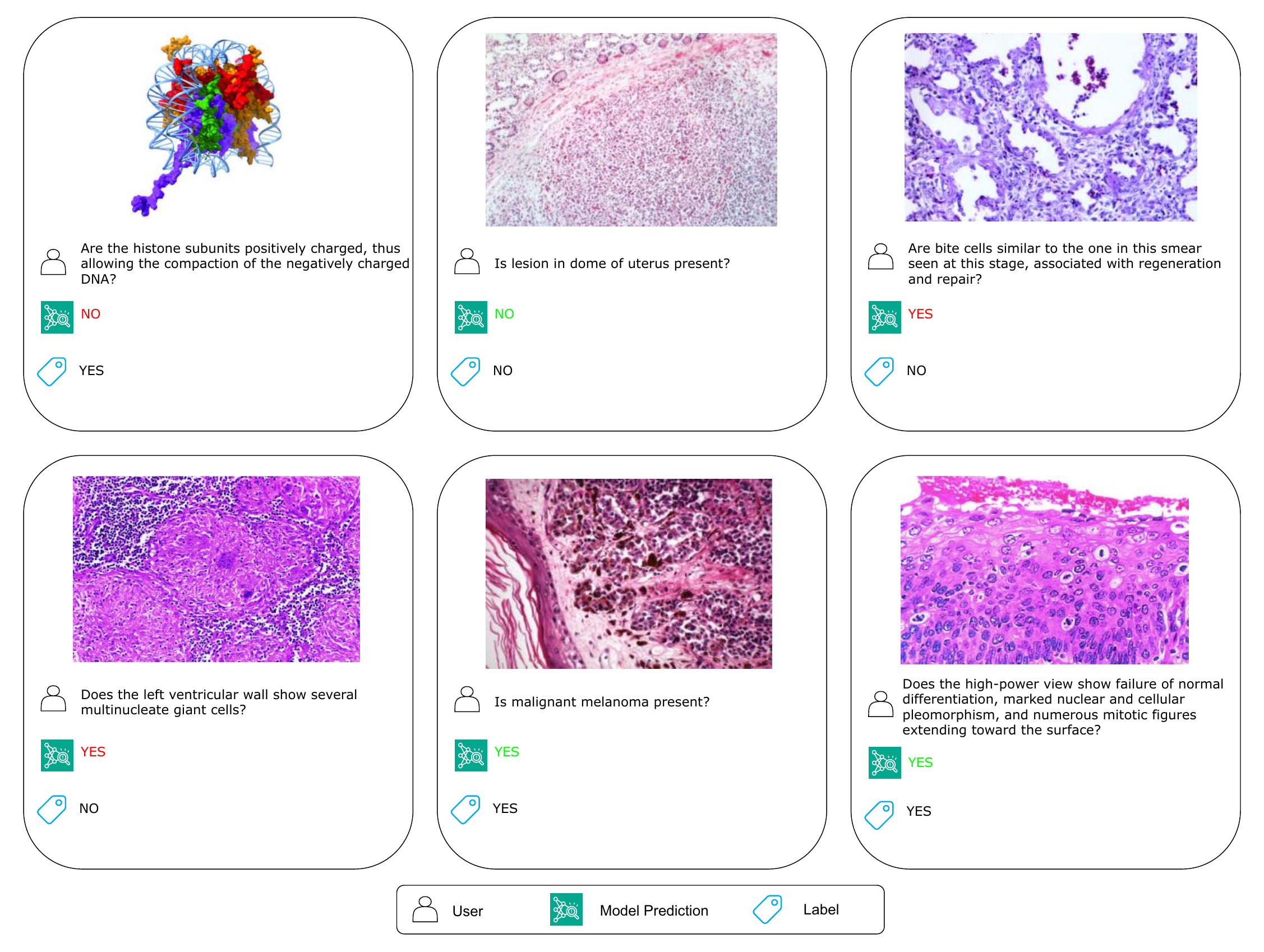}

    \caption{
Qualitative examples from the centralized PathVQA baseline on close-ended VQA task.}
    \label{fig:pathvqa_qual}
\end{figure*}

To provide a clearer sense of the underlying tasks, we present qualitative predictions from centralized baseline models on PathVQA (\cref{fig:pathvqa_qual}) and EHRXQA (\cref{fig:ehrxqa_qual}). These examples illustrate the nature of the multimodal reasoning required in both datasets. Note that these examples are intended solely to demonstrate the task characteristics rather than to compare different federated training methods.
Aside from these examples, we also include qualitative results for Fed-BraTS-GLI2024, showcasing the input MRI modalities (T2-FLAIR, T1Gd, T1, and T2), the corresponding ground-truth tumor segmentation mask, and the predicted masks produced by six federated learning algorithms. These visualizations are drawn from the synthetic non-IID setting with Dirichlet $\alpha = 0.2$ across three clients, and are provided to offer a representative view of the segmentation outputs across methods.

\begin{figure*}[ht]
    \centering
    \includegraphics[width=\linewidth]{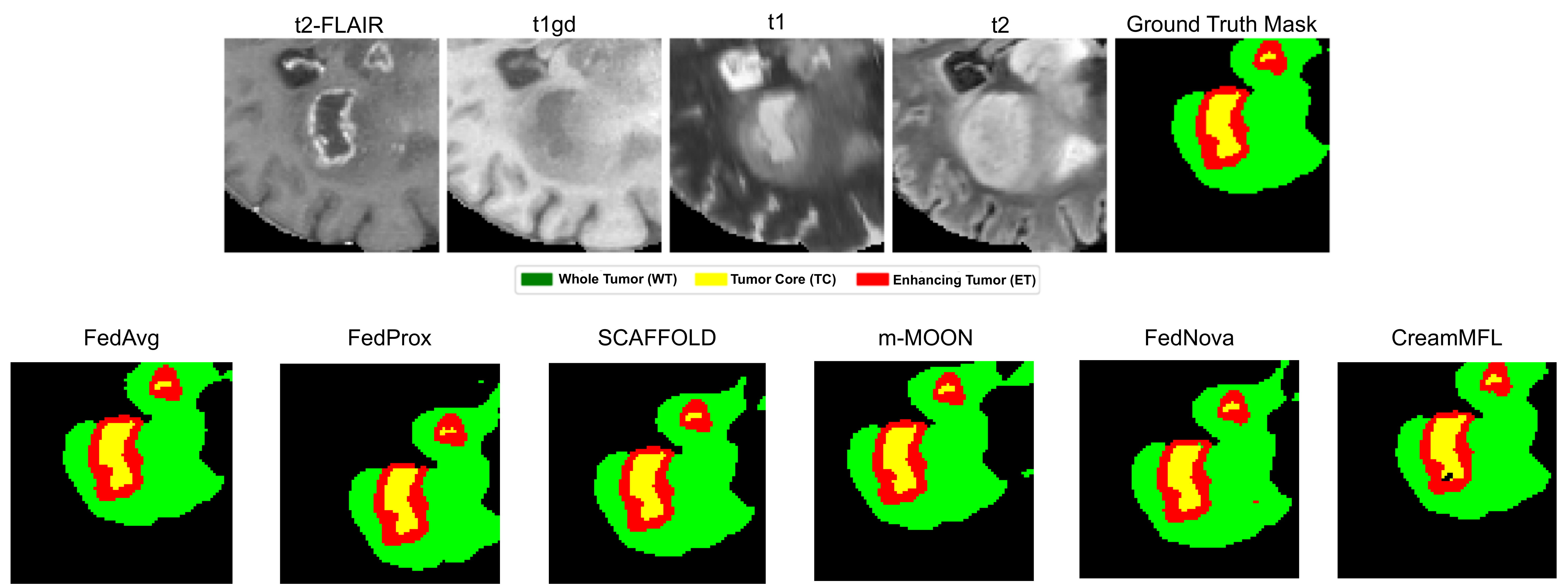}
    \caption{Multimodal Brain Tumor Segmentation results on Fed-BraTS-GLI2024. Top row shows the four MRI input modalities (T2-FLAIR, T1Gd, T1, T2), followed by the ground-truth segmentation mask , and the bottom row shows the predictions from six federated learning algorithms under the synthetic non-IID setting ($\alpha = 0.2$, three clients).}
\end{figure*}
% \section{Convergence Curves}
% \label{sec:supp_convergence}